\documentclass{article}

\usepackage[accepted]{icml2024}
\usepackage{microtype}
\usepackage{graphicx}
\usepackage{subfigure}
\usepackage{booktabs} 

\usepackage{algorithm}
\usepackage{algorithmic}

\usepackage{amsmath}
\usepackage{amssymb}
\usepackage{mathtools}
\usepackage{amsthm}

\DeclareMathOperator*{\argmin}{arg\,min}

\usepackage[capitalize,noabbrev]{cleveref}

\theoremstyle{plain}
\newtheorem{thm}{Theorem}[section]
\newtheorem{prop}[thm]{Proposition}
\newtheorem{lemma}[thm]{Lemma}

\theoremstyle{definition}
\newtheorem{defn}[thm]{Definition}

\theoremstyle{remark}

\usepackage[textsize=tiny]{todonotes}

\icmltitlerunning{Multi-Bit Distortion-Free Watermarking for LLMs}

\usepackage{float}
\usepackage{comment}
\usepackage{tikz}
\usetikzlibrary{patterns,decorations.pathreplacing}
\usepackage{xcolor}
\usepackage{bigints}
\usepackage{bbm}
\usepackage{dblfloatfix}
\usepackage{pifont}%

\renewcommand{\Pr}{\mathsf{P}}
\newcommand{\Ex}{\mathsf{E}}
\newcommand{\Var}{\mathsf{Var}}

\newcommand{\PROMPT}{\alpha}

\newcommand{\bw}{\boldsymbol{w}}
\newcommand{\bx}{\boldsymbol{x}}
\newcommand{\by}{\boldsymbol{y}}
\newcommand{\bz}{\boldsymbol{z}}
\newcommand{\bW}{\boldsymbol{W}}
\newcommand{\bY}{\boldsymbol{Y}}
\newcommand{\bL}{\boldsymbol{L}}
\newcommand{\Uniform}{\mathsf{Uniform}}
\newcommand{\Er}{\mathrm{Er}}

\begin{document}

\twocolumn[
\icmltitle{Multi-Bit Distortion-Free Watermarking
for Large Language Models}



\icmlsetsymbol{equal}{*}

\begin{icmlauthorlist}
\icmlauthor{Massieh Kordi Boroujeny}{ECE}
\icmlauthor{Ya Jiang}{CS}
\icmlauthor{Kai Zeng}{ECE}
\icmlauthor{Brian Mark}{ECE}

\icmlaffiliation{ECE}{Department of Electrical and Computer Engineering, George Mason University, Fairfax, VA, USA}
\icmlaffiliation{CS}{Department of Computer Science, George Mason University, Fairfax, VA, USA}

\icmlcorrespondingauthor{Massieh Kordi Boroujeny}{mkordibo@gmu.edu}
\end{icmlauthorlist}

\icmlkeywords{LLM Watermarking, Distortion-free, Multi-bit Watermark}

\vskip 0.3in
]



\printAffiliationsAndNotice{}  

\begin{abstract}
Methods for watermarking large language models have been proposed that distinguish AI-generated text from human-generated text by slightly altering the model output distribution, but they also distort the quality of the text, exposing the watermark to adversarial detection. More recently,  {\em distortion-free} watermarking methods were proposed that require a secret key to detect the watermark. The prior methods generally embed {\em zero-bit} watermarks that  do not provide additional information beyond tagging a text as being AI-generated.  We extend an existing zero-bit distortion-free watermarking method by embedding {\em multiple bits} of meta-information as part of the watermark. We also develop a computationally efficient decoder that extracts the embedded information from the watermark with low bit error rate.


\end{abstract}

\section{Introduction}

 

The emergence of large language models (LLMs) has ushered in 
the dawn of general artificial intelligence \cite{dong2023towards, tamkin2021understanding, brown2020language, touvron2023llama, zhang2022opt, chowdhery2022palm, ye2023natural} due to their remarkable capabilities in understanding and generating human-like text. LLMs can be fine-tuned for a wide range of natural language processing tasks, such as text summarization, translation, question-answering, and more.  Their versatility allows for the development of more generalized AI systems \cite{wu2023next, ge2023openagi, mialon2023augmented, li2021prefix}. The natural language generation capabilities of these models contribute to more natural and human-like interactions between machines and humans, making AI systems more accessible and user-friendly \cite{chan2023chateval, shen2023hugginggpt, wu2023visual, vemprala2023chatgpt}. 

Unfortunately, along with these attractive features, LLMs also create opportunities for malicious use, such as misinformation spreading, inappropriate content generation, unethical activities engagements, 
academic cheating, etc. \cite{yang2023anatomy, lapid2023open, bommasani2021opportunities, goldstein2023generative}.
To bolster AI accountability, it is crucial to be able to ascertain the provenance of a given text, i.e., whether the text was generated by an LLM or crafted by a human, and if the text was generated by an LLM, which LLM was
used.

Post-hoc detectors, which were proposed as an initial approach to address this concern, are based on the use of statistical outliers~\cite{lavergne2008detecting,beresneva2016computer, gptzero, mitchell2023detectgpt} or training a binary classifier over the human-generated and LLM-generated texts~\cite{gpt2,bakhtin2019real,fagni2021tweepfake}. However, these methods tend to be rendered ineffective as LLM-generated texts have become increasingly similar to the human-generated texts with advances in LLMs. 

Another approach to AI accountability is to embed a watermark in a generated text. Watermarks are embedded by means of intentional and hopefully imperceptible modifications to an LLM. Recently, Kirchenbauer et al. \cite{Kirchenbauer2023} presented the first watermarking scheme for LLMs.
However, in their watermarking scheme the distribution of the generated watermarked text deviates from that of the original LLM distribution, which results in distortion of the text quality.

A good watermarking scheme should satisfy the following:
\renewcommand{\labelenumi}{\Roman{enumi}.}
\begin{enumerate}
    \item \textbf{\textit{Distortion-free}}: The watermarked text should have the same output distribution as the original LLM. 
    
    \item \textit{Low probability of false alarm}: Human-generated text should be detected as AI-generated with negligible probability.
    
    \item \textit{High probability of correct detection}:  AI-generated text should be detected as such
     with high probability.
     
\end{enumerate}
To achieve these desirable properties, watermarking schemes were further developed in \cite{Aaronson2023, Christ2023, kuditipudi2023robust}. 
\cite{Christ2023} proposed a watermarking scheme that 
claims to have properties I--III. 
Their method relies on binarization of the token set and token generation based on the value of a pseudorandom function (PRF). \cite{Aaronson2023} proposed a similar method, which satisfies the above three properties but 
does not involve binarization.

The above-mentioned references are considered zero-bit watermarking schemes in the sense that the watermark does not embed any information beyond differentiating an AI-generated text from a human-generated one. 
In practice, it is crucial to encode meta-information such as the language model name, model version, and generation time within the watermark. For example, encoding meta-information in the watermark supports forensic analysis in case of misuse. It helps trace back the origin of content and assists in determining whether a specific model or version was involved.  Moreover, the incorporation of meta-information in the watermark aligns with the need for responsible and accountable AI practices.

Therefore, we extend the above list with the following additional properties:
\begin{enumerate}
    \setcounter{enumi}{3}
    \item \textbf{\textit{Multi-bit embedding}}:  The watermark encodes multiple
      bits of meta-information.
   
    \item \textit{High probability of correct decoding}: The bit error rate (BER) in decoding the embedded information bits should be low.
    
    \item \textit{Efficient decoding}: The decoding algorithm should be efficient and not require exhaustive search over all the possible embedded information bits. 
\end{enumerate}
In this paper, we develop the first efficient multi-bit distortion-free watermarking scheme that satisfies all of the above properties I--VI. Table~\ref{table:comparison} compares our proposed scheme with the state-of-the-art approaches.

\begin{table}
\centering
\caption{Comparison between other LLM watermarking schemes
 and this work.
 \\
{\bf A}: \cite{Kirchenbauer2023}, \cite{Kirchenbauer2023a},\cite{liu2023semantic};
{\bf B:} \cite{Christ2023},\cite{Aaronson2023},\cite{kuditipudi2023robust};
{\bf C:} \cite{Wang2023a}, \cite{abdelnabi2021adversarial}, \cite{yoo2023robust}, \cite{fernandez2023bricks}. }
\begin{tabular}{|l|l|l|l|c|}\hline
    & \textbf{A} & \textbf{B} & \textbf{C}  & \textbf{This work}    
\\
\hline
high detection rate  & \ding{51}   & \ding{51}  & \ding{51} & \ding{51} 
\\
\hline
low false alarm   & \ding{51}  & \ding{51}  & \ding{51}  & \ding{51} 
\\
\hline
\textbf{distortion-free} & \textcolor{red}{\ding{55}} & \textcolor{green}{\ding{51}}  &  \textcolor{red}{\ding{55}}  & \textcolor{green}{\ding{51}} 
\\
\hline
\textbf{multi-bit watermark} & \textcolor{red}{\ding{55}} & \textcolor{red}{\ding{55}} & \textcolor{green}{\ding{51}} & \textcolor{green}{\ding{51}} 
\\
\hline
low BER  & \ding{55}  & \ding{55} & \ding{55}   & \ding{51} 
\\
\hline
efficient decoding & \ding{55}  & \ding{55} & \ding{55}   & \ding{51} 
\\
\hline
\end{tabular}
\label{table:comparison}
\end{table}

The remainder of the paper is organized as follows.  In Section~\ref{sec:prelim}, we show how distortion-free watermarking schemes can be represented by a distortion-free mapping rule.  In Section~\ref{sec:christ.}, we apply this generalization to the scheme of \cite{Christ2023} and correct a flaw in their design of the decoder.  In Section~\ref{sec:watermarkWcoding}, we 
extend the distortion-free mapping rule
to a {\em multi-bit} distortion-free mapping rule and then develop a 
new multi-bit distortion-free watermarking algorithm. Section~\ref{sec:expr} presents numerical results from our simulation study.  Concluding remarks are given in Section~\ref{sec:con}.

\section{Preliminaries}
\label{sec:prelim}

Random variables are denoted by boldfaced letters, e.g., $\mathbf{w}$, whereas, specific realizations of a random variable are denoted by non-boldfaced letters, e.g., $w$. In this case, $\mathbf{w} = w$ means that
the random variable $\mathbf{w}$ gets value $w$. For brevity, in conditional probability expressions, we abbreviate expressions such as $\mathbf{w} = w$ by simply $w$. We use uppercase letters to denote vectors and calligraphic font to denote sets. We use $X_{[i:j]} := (x_i, x_{i+1}, \ldots, x_j)$ and
$X_{[n]} := (x_1, \ldots, x_n)$ to denote subsequences
of a  sequence $\{ x_n \}$. 

\begin{defn}[\bf{Language Model}]
A {\em vocabulary} $\mathcal{V} = \{v_1, v_2, 
\ldots, v_{|\mathcal{V}|}\}$ is a set of tokens.
Given a generated sequence of tokens $W_{[t-1]} \in \mathcal{V}^{t-1}$ and a 
prompt $\alpha = W_{[-(N_p-1):0]} \in \mathcal{V}^{N_p}$, where $N_p$ is the
length of the prompt, a {\em Language Model} 
$\mathbb{M}$ is specified by a conditional distribution
\begin{align}
    p_{t,i} &= p_{\mathbb{M}} (v_i \mid W_{[t-1]}, \PROMPT) \nonumber \\
    &:= 
\Pr_{\mathbb{M}} \left \{ \bw_t = v_i \mid 
        \bW_{[t-1]} = W_{[t-1]}, \PROMPT  \right \}, 
\label{eq:llm_distr}
\end{align}
where $i= 1, \ldots, |\mathcal{V}|$
for the $t^{th}$ token $\bw_t \in \mathcal{V}$.
We use $D_t =(p_{t,1}, p_{t,2}, \ldots, p_{t,|\mathcal{V}|})$ to denote this conditional probability distribution over the set $\mathcal{V}$
and define $D_t[v_j] = p_{t,j}$, for $j =1,\ldots,|\mathcal{V}|$.
\label{defn:LLM}
\end{defn}
Note that $\bw_t$ is a nonstationary discrete-time random process with
conditional probability distribution given by \eqref{eq:llm_distr}.
The response of a language model $\mathbb{M}$ to a prompt $\PROMPT$ is a random
vector $\mathbb{M}(\PROMPT) := \bW_{[\bL]}$, where $\bL$ is a random variable,
with conditional distribution given by  
\begin{align}
\Pr & \left \{ \mathbb{M}(\PROMPT)= W_{[\ell]} \right \} =  
 \prod_{t=1}^\mathbf{\ell} p_{\mathbb{M}}( w_t \mid 
  W_{[t-1]}, \PROMPT ) \nonumber \\
 &=p_{\mathbb{M}}( W_{[\ell-1]} \mid \PROMPT )
 \cdot p_{\mathbb{M}}(w_{\ell} \mid W_{[\ell-1]}, \PROMPT ), ~~\ell = 1, 2, \ldots,
 \label{eq:conditionDist}
\end{align} 
where $p_{\mathbb{M}}( W_{[0]} \mid \PROMPT ) = 1$, and $p_{\mathbb{M}}(\cdot \mid W_{[0]}, \PROMPT )= p_{\mathbb{M}}(\cdot \mid  \PROMPT )$. This random response is generated by sampling from $D_t$ until a special terminating token $\mathsf{done} \in \mathcal{V}$ is generated. 
Therefore, $\mathbb{M}(\PROMPT) = W_{[L]}$ implies that $w_{[L]} = \mathsf{done}$.
When we talk about altering a language model, we mean $D_t \rightarrow D'_t$, where the prompt $\PROMPT$ and the 
sequence generated tokens so far, i.e., $W_{[t-1]}$, are fixed. 
The entropy of the response of a language model $\mathbb{M}$ to a prompt 
$\PROMPT$ is defined as 
\begin{equation}
    H( \PROMPT ) := \Ex_{\mathbb{M}(\PROMPT)}
    \{- \ln\Pr\{\mathbb{M}(\PROMPT) \} \} 
    \label{eq:entropy}
\end{equation}
For further discussion of $H(\PROMPT)$ see Appendix~\ref{subsec:entropy_appendix}.

Having established a language model definition, we now define a distortion-free watermarking algorithm.
\begin{defn}
    \label{defn:distortion-free}
    A watermarking algorithm is {\em distortion-free} if for any prompt
    $\PROMPT$ and a sequence of watermarked generated text $W_{[\ell]}$, we have 
    \begin{align*}
        \Pr\{ & \bw_t = w_t  \mid \bW_{[t-1]} = W_{[t-1]}, \PROMPT \}    \nonumber \\
        &= p_{\mathbb{M}}( w_t \mid W_{[t-1]}; \PROMPT ) = D_t[w_t],
    \end{align*}
    for all $t \in \{1,\ldots,  \ell \}$.
\end{defn}
In other words, a watermarking algorithm is distortion-free if the watermarked text has the same distribution $D_t$ as the non-watermarked text.

Following~\cite{Christ2023}, our proposed watermarking algorithms 
use pseudorandom functions (PRFs) to generate random numbers. PRFs are defined as follows, 
\begin{defn}[\bf{PseudoRandom Functions (PRF)}]
Let $\mathcal{F} = \{F_{\mathsf{sk}}: \{0,1\}^{l_1(\lambda)} \rightarrow \{0,1\}^{l_2(\lambda)} | \mathsf{sk} \in \{0,1\}^\lambda \}$ be a family of functions. $\mathcal{F}$ is a PRF if $F_{\mathsf{sk}}$ is efficiently computable and for all probabilistic polynomial-time distinguishers $D$, 
\begin{align*}
    \bigg\lvert\Pr_{\mathsf{sk} \leftarrow \{0,1\}^\lambda}\lbrace D^{F_{\mathsf{sk}}(\cdot)}(1^\lambda) = 1  \rbrace &- \Pr_f\lbrace D^{f(\cdot)}(1^\lambda) = 1  \rbrace \bigg\rvert 
    \\
    &\leq \mathsf{negl}(\lambda),
\end{align*}
where $f: \{0,1\}^{l_1(\lambda)} \rightarrow \{0,1\}^{l_2(\lambda)}$ denotes a random function. A function $g(\lambda)$ is negligible,
denoted by $\mathsf{negl}(\lambda)$, if $g(\lambda) \in O \left (\frac{1}{\mathsf{poly}(\lambda)} \right )$ for every $\mathsf{poly}(\cdot)$. 
\label{defn:PRF}
\end{defn}

The basic approach to distortion-free watermarking
is to embed the watermark in the correlation between $\{ \bw_t \}_{t=1}^\ell$ and a specially  crafted i.i.d.\ sequence $\{ \by_t \}_{t=1}^\ell$~\cite{kuditipudi2023robust, Christ2023, Aaronson2023}, where $\ell$ is the length of the generated text. In~\cite{Christ2023,Kirchenbauer2023}, $\by_t \sim \Uniform[0,1]$.
In the watermarking process, the token $\bw_t$ is generated such that it has 
high correlation with $\by_t$, for $t = 1,\ldots, \ell$.
On the other hand, for the human-generated text or non-watermarked text, the token $\bw_t$ is independent of $\by_t$, for $t = 1,\ldots, \ell$.
For a given sequence of tokens $W = W_{[\ell]}$ and a sequence of generated random numbers 
$Y = Y_{[\ell]}$, the detection algorithm is a statistical test $\psi(W, Y)$ 
that checks whether the correlation between the two sequences exceeds a threshold.

In the watermarking process, the correlation between $\{ \bw_t \}$ and $\{ \by_t \}$ is introduced using a watermarking mapping rule. 
\begin{defn}[\bf{Watermarking mapping rule}]
    For a random variable $\by$ defined over the sample space $\Omega$ with distribution $\Pr_{\by}$, a watermarking mapping rule $\Gamma(\Omega, \mathcal{V}; \Pr_{\by})$ is a process of mapping a partition of a sample space $\Omega$ into the token set $\mathcal{V}$ of the language model $\mathbb{M}$. This process is done in two steps:  First, a partition $U=\{A_1,\ldots, A_{|\mathcal{V}|}\}$ on the sample space $\Omega$ is formed. Then each partition part $A_j$ is mapped to $v_j$, for $j=1,\ldots,|\mathcal{V}|$ (see Appendix~\ref{subsec:watermarking_mapping_appendix}).
    \label{def:mappingRule}
\end{defn}

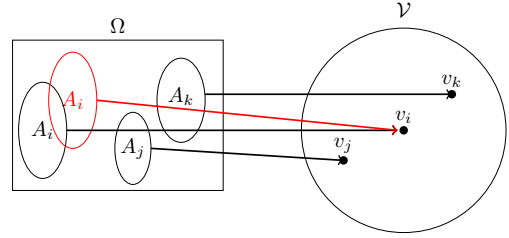
\begin{figure}[t]
  \centering
  \scalebox{0.8}{
  \begin{tikzpicture}
    \draw (0, 0) rectangle (3.5, 2.5);
    \node at (0.5, 1) {$A_i$};
    \node at (2, 0.7) {$A_j$};
    \node[above] at (2.8, 1.3) {$A_k$};
    \node[red]at (1, 1.5) {$A_i$};
    \node [above]at (1.75, 2.5) {$\Omega$};

    \draw (0.5,1) ellipse (0.4cm and 0.8cm);
    \draw (2,0.7) ellipse (0.3cm and 0.6cm);
    \draw (2.8,1.5) ellipse (0.4cm and 0.7cm);
    \draw [red](1,1.5) ellipse (0.4cm and 0.8cm);

    \draw (6.5, 1) circle (1.7cm);
    \node at (6.5, 3) {$\mathcal{V}$};
    \fill (6.5, 1) circle (2pt) node[above] {$v_i$};
    \fill (5.5, 0.5) circle (2pt) node[above] {$v_j$};
    \fill (7.3, 1.6) circle (2pt) node[above] {$v_k$};
    
    \draw[->, thick] (0.9, 1) -- (6.4, 1);
    \draw[->, thick] (2.3, 0.7) -- (5.5, 0.5);
    \draw[->, thick] (3.2, 1.6) -- (7.3, 1.6);
    \draw[->, thick, red] (1.4, 1.5) -- (6.4, 1);
  \end{tikzpicture}
  }
  \caption{Watermarking mapping rule $\Gamma(\Omega, \mathcal{V})$.} 
    \label{fig:distotionFree}
\end{figure}

The distortion-free property of a watermarking algorithm following a watermarking mapping rule $\Gamma_t(\Omega, \mathcal{V}; \Pr_{\by})$ is characterized as follows. 
\begin{prop}
A watermarking algorithm following a watermarking mapping rule $\Gamma_t(\Omega, \mathcal{V}; \Pr_{\by})$ is distortion-free 
if and only if for every prompt $\PROMPT$ and the past generated tokens $W_{[t-1]}$, 
\begin{align}
    \Pr\{\by_t \in A_{j,t}\} &= 
     p_{\mathbb{M}}(v_j \mid W_{[t-1]}, \PROMPT ) =D_t[v_j] .
\end{align}
\label{prop:distortion-free}
\end{prop}

\section{Zero-bit Distortion-free Watermarking}
\label{sec:christ.}

In this section, we review the zero-bit distortion-free watermarking scheme of \cite{Christ2023} and address a problematic issue with the statistical test used in their scheme.  We then propose a modification to the random initialization in
this method and develop an alternative zero-bit distortion-free watermarking algorithm, which we extend further
to a multi-bit algorithm in Section~\ref{sec:watermarkWcoding}.

\subsection{Binarization of language model}

Watermarking in~\cite{Christ2023} follows a watermarking mapping rule and is done on binary tokens. Language model $\mathbb{M}$ with token set $\mathcal{V}$ is converted into a language model $\mathbb{M}^{b}$ with binary token set  $\mathcal{V}^b=\{0,1\}$.
This conversion is done as follows: First, each token $v \in \mathcal{V}$ is represented as a distinct binary string in $\{0,1\}^{\log|\mathcal{V}|}$,
where $\log(\cdot)$ denotes the base-2 logarithm. 
In this way, sampling $\log|\mathcal{V}|$ times from the language model $\mathbb{M}^{b}$ is equivalent to sampling one token from language model $\mathbb{M}$.
Henceforth, we will assume binarization has been applied (see Appendix~\ref{subsec:Binarization_appendix} for further details).

\subsection{Watermarking without random initialization}
\label{subsec:christWORandomChunk}

\begin{figure}
    \centering
    \scalebox{0.8}{
    \begin{tikzpicture}
        \draw[thick] (-0,0) -- (5,0);
        \filldraw[black] (0,0) circle (2pt) node[anchor=north]{0};
        \filldraw[black] (5,0) circle (2pt) node[anchor=north]{1};
        \filldraw[black] (3,0) circle (2pt) node[anchor=north]{$p_i(1)$};
        \node [below] at (1.5,-0.5) {$0 \leq \by_i \leq p_i(1)$};
        \node [below] at (4,-0.5) {$p_i(1) \leq \by_i \leq 1$};
        \draw [->, thick] (1.5, 0.3) [out=30,in=150] to (7, 0);
        \draw [->, thick] (4, 0.3) [out=30,in=150] to (6, 0);
        \filldraw[black] (6.1,0) circle (2pt) node[anchor=north]{0};
        \filldraw[black] (7.1,0) circle (2pt) node[anchor=north]{1};
        \draw (6.5, 0) circle (1cm);
        \node [above] at (6.5, 1) {$\mathcal{V}^b$};
        \draw [decorate,decoration = {brace}, thick] (0.2,0.1) --  (2.8,0.1);
        \draw [decorate,decoration = {brace}, thick] (3.2,0.1) --  (4.8,0.1);
    \end{tikzpicture}
    }
    \caption{Watermarking mapping rule in~\cite{Christ2023}.} 
    \label{fig:chrirstMappingRule}
\end{figure}
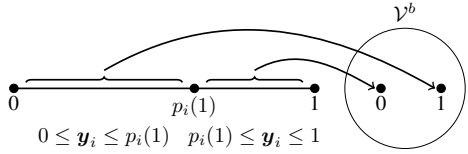

The watermarking algorithm decides the value for each binary token
according to a watermarking mapping rule as shown in Figure~\ref{fig:chrirstMappingRule}.

Given a prompt~$\PROMPT^b$ and the past generated tokens $W^b_{[i-1]}$, the watermarking mapping rule $\Gamma_i(\Omega, \mathcal{V}^b; \Pr_{\mathbf{y}})$ is specified
as follows: 
\begin{align}
&\Omega = [0 , 1], \quad \quad \by_i\sim \Uniform[0,1],\nonumber \\
&\Gamma_i(\Omega) = \{A_{1,i} = [p_i(1), 1], A_{2,i} = [0, p_i(1)) \}, \nonumber \\
&\Gamma_i(A_{1,i}) = 0, \quad \Gamma_i(A_{2,i}) = 1,
\label{eq:christMappingRule}
\end{align}
where $p_i(1)$ is the probability of the $i$-th token being 1 according to $\mathbb{M}^{b}$.
According to~\eqref{eq:christMappingRule},
\begin{align*}
    \Pr\{\by_i\in A_{1,i}\} = p_i(0),~
    \Pr\{\by_i \in A_{2,i}\} = p_i(1).
\end{align*}
By Proposition~\ref{prop:distortion-free}, a watermarking scheme following this watermarking mapping rule is distortion-free.
The random variable $Y_i$ can be generated using a PRF $F_{\mathsf{sk}}: \{0,1\}^{\mathsf{poly}_1(\lambda)}\rightarrow\{0,1\}^{\mathsf{poly}_2(\lambda)}$, with a secret key $\mathsf{sk} \in \{0,1\}^\lambda$, shared between the encoder and the decoder. 
Here, $\lambda$ is the security parameter of the watermarking algorithm (see~\cite{Christ2023}).

Unlike~\cite{Christ2023}, we will not use the index $i$ as the the input to this PRF, as this choice of input will make the watermark very prone to a simple deletion attack, and removing one word in a watermarked text can render the detection process useless. 
Instead we use the \textit{context}
ngram $S_{i,h} = W^b_{[i-h:i-1]}$ as the input to the PRF for the $i$-th token.
Usually $h$ is chosen such that $\lfloor h/\log|\mathcal{V}| \rfloor\leq 8$. In our work, we set $\lfloor h /\log|\mathcal{V}|\rfloor= 5$.
Here, $\mathsf{poly}_1$ is chosen such that the ngram $S_{i,h}$ is not too long for the PRF, and if it is too short it will be padded. 
On the other hand, let $z$ be the integer representation of the output of PRF; then taking $\frac{z}{2^{\mathsf{poly}_2(\lambda)}}$ results in a real number in $[0,1]$. 
In our notation, we assume these steps are included and $F_{\mathsf{sk}}(i) \in [0,1]$. 
The watermarking encoding scheme can be summarized as follows:
\begin{align}
\by_i = F_{\mathsf{sk}}(S_{i,h}), ~~\bw^b_{i}=\left\{\begin{array}{ll}
     1, &  0 \leq \by_i < p_i(1),\\
     0,  &  p_i(1) \leq \by_i \leq 1 .
 \end{array}\right. 
 \label{eq:watermarkingRuleChrist}
\end{align}

For $W^b = W^b_{[\ell]}$ and $Y = Y_{[\ell]}$, the decoder uses a statistical test $\psi(W^b, Y)$ to check whether the correlation between $Y$ and $W^b$ exceeds a certain threshold. 
The null hypothesis is $\mathcal{H}_0:$ {\em text is non-watermarked} while
the alternative hypothesis is $\mathcal{H}_1:$ {\em text is watermarked}. 
First, for each binary token a score value is calculated. 
This score value depends on the binary token value, $w^b$, and the uniform random number generated, $y$. 
Define $C(W^b, Y)$
as the sum of the score values for all $(w^b_i,y_i)$, $i= 1, \ldots, \ell$:
\begin{equation}
  C(W^b, Y) := \sum_{i=1}^\ell s(w^b_i, y_i) .
  \label{eq:CWY}
\end{equation}
The $p$-value for the observed value $z$ of a random variable $\bz$ is defined as the probability of observing a value at least as extreme as the observed value $z$ under $\mathcal{H}_0$, i.e.,
\begin{equation}
 p\text{-value}(z) = \Pr\{\bz > z \mid \mathcal{H}_0\}. 
 \label{eq:pvalue}
\end{equation}
The $p$-value for $C(W^b, Y)$ is compared to a threshold FPR, where FPR is the maximum tolerable false positive rate for detecting a non-watermarked text as watermarked. 
If $p$-value$(C(W^b, Y)) \leq\text{FPR}$, the text is detected as watermarked (cf. Algorithm~4 in \cite{Christ2023}). 
This leads to a critical region $D_c = \{C(W^b, Y) \geq \theta\}$, such that $\mathcal{H}_0$ is rejected if $C(W^b, Y) \in D_c$, and a region of acceptance $D^c_c = \{C(W^b, Y) < \theta\}$, where $\mathcal{H}_0$ is accepted if $C(W^b, Y) \in D^c_c$.
In other words, for a non-watermarked text $W_{\mathsf{NW}}^b$ and the constructed $\mathbf{Y}$, the threshold $\theta$ is chosen such that, $\Pr\{C(W_{\mathsf{NW}}^b, \mathbf{Y}) \geq \theta\} \leq \text{FPR}$, for all $W_{\mathsf{NW}}^b$ with length $\ell$. 
This choice of threshold $\theta$, on the other hand, affects the false negative rate.
If in addition to FPR, there is also a maximum tolerable false negative rate FNR, this bound will yield a minimum length of watermarked text such that both false positive rate and false negative rate are bounded by FNR and FPR, respectively (see Appendix~\ref{subsubsec:appendix_WOr_1}). 

As in~\cite{Christ2023}, the score function is calculated as follows,
\begin{equation}   
s(w^b_{i},y_i)=\left\{
\begin{array}{ll}
\ln\frac{1}{y_i}, & w^b_{i}=1,\\
\ln\frac{1}{1-y_i}, & w^b_{i}=0.
\end{array}
\right.
\label{eq:scoreChrist}
\end{equation}

This score function is designed such that, given a prompt~$\PROMPT^b$ and the past generated tokens $W^b_{[i-1]}$, the expected value of the score function for the $i$-th token, if it is watermarked, is greater than its expected value if it is non-watermarked (see Appendix~\ref{subsubsec:appendix_WOr_2}). 

Using the central limit theorem (CLT), we can approximate the distribution of 
$C(W_{\mathsf{W}}^b, Y)$ (see Appendix~\ref{subsubsec:appendix_WOr_3}). 
There can be correlation between the $s(w_i^b, y_i)$'s. 
This correlation can stem from the correlation between the tokens generated by an LLM or from similar ngrams with length $h$ appearing throughout the text. 
Not much can be done for the first cause of correlation, however, by adopting the idea from~\cite{fernandez2023bricks}, in the detection process we can eliminate the second cause. Namely, we consider any ngram ``context plus current token,'' i.e., 
$w^b_{[i-h:i]}$, only once. 
In other words, if for different values of $i$, the corresponding context plus current token is repeated before, we will not include its corresponding score value in $C(W^b, Y)$.
Therefore, to simplify our derivation, in applying the CLT we shall assume that
the $s(\bw_i^b, \by_i)$'s are independent. 
Our experimental studies have shown that this assumption does not adversely affect
the detection algorithm. 

The threshold $\theta$ and the critical region $D_c$ are chosen such that for a non-watermarked text $W^b_{\mathsf{NW}}$ and the constructed $Y$, $\Pr\{C(W_{\mathsf{NW}}^b, \mathbf{Y}) > \theta\} \leq \text{FPR}$. Using~\eqref{eq:sumScorePDFNonWatermarkedErlang}, for a given $L$ and FPR, we can derive $\theta$ as follows,
\begin{equation}
    \theta = \bar{F}^{-1}_{E_L(1)}(\text{FPR}) = Q^{-1}(L,\text{FPR}),
    \label{eq:thetaFPR}
\end{equation}
 where $\bar{F}_{\mathrm{Er}_L(1)}(x)$ is the tail distribution function of a random variable $\bx \sim \mathrm{Er}_L(1)$
 and $\mathrm{Er}_L(\lambda)$ denotes the Erlang distribution with shape parameter~$k$ and rate~$\lambda$:
\begin{equation}
    \bar{F}_{\mathrm{Er}_L(1)}(x) =\Pr\{\bx > x\} =\frac{\Gamma(L,x)}{\Gamma(L)} = Q(L,x),
    \label{eq:FbarErlang}
\end{equation}
 where $\Gamma(L,x)$ is the upper incomplete gamma function and $Q(L,x)$ is the regularized gamma function
 (see Appendix~\ref{subsubsec:appendix_WOr_3}).
 
We have derived an approximation for $L_{\min}$ 
(see Appendix~\ref{subsubsec:appendix_WOr_4}):
\begin{align}
 L_{\min}  \approx  \frac{f_1^2(\text{FPR},\text{FNR})}{\zeta^b(\PROMPT^b)^2} 
    \label{eq:LminApprox}
\end{align}
where
\begin{align}
 f_1(\text{FPR},\text{FNR}) = \sqrt{2\ln{\frac{1}{2\text{FPR}}}}+\sqrt{4.4\ln{\frac{1}{2\text{FNR}}}}
 \label{eq:f1}
\end{align}
The approximation in~\eqref{eq:LminApprox} is similar to the lower bound approximation in~\cite{Aaronson2023}, i.e., $O(\frac{1}{\zeta(\PROMPT)^2}\ln\frac{1}{\eta})$ with FPR=FNR$=\eta$. 
Using~\eqref{eq:zeta&zetab}, we can express $L_{\min}$ in~\eqref{eq:LminApprox} in terms of the average conditional entropy, conditioned on the past tokens, per token of response of the language model $\mathbb{M}$ to the prompt $\PROMPT$, i.e., $\zeta(\PROMPT)$ as 
\begin{equation}
   \log|\mathcal{V}| \frac{f_1^2(\text{FPR},\text{FNR})}{\zeta(\PROMPT)^2} \approx \frac{L_{\min}}{\log|\mathcal{V}|}.
     \label{eq:LlowerBound}
 \end{equation}
 
\begin{figure}
\centering
\includegraphics[scale = 0.45]{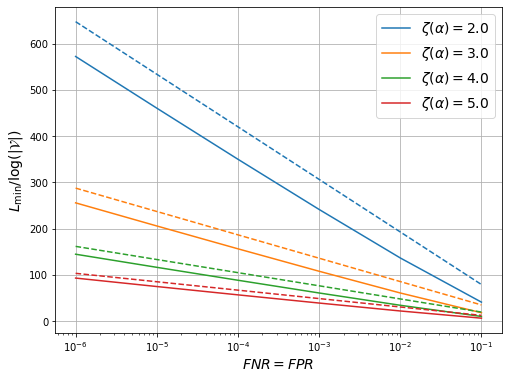}
\caption{Exact and approximate (dashed lines) $L_{\min}$, $|\mathcal{V}|= 50272$.}
\label{fig:Lmin}
\end{figure}

 Based on~\eqref{eq:LlowerBound}, the number of required tokens to achieve a desired probability of error for a watermarking scheme, on a binarized language model $\mathbb{M}^b$, is $\log|\mathcal{V}|$ times than what it would have been
 under $\mathbb{M}$.
The exact and approximate values of $L_{\min}$ derived by the numerical method mentioned here and by using~\eqref{eq:LminApprox}, for different values of $\zeta(\PROMPT^b)$ for vocabulary size $|\mathcal{V}|= 50272$, are shown in Figure~\ref{fig:Lmin}, where the approximate values are depicted by dashed lines. In Figure~\ref{fig:Lmin}, the bounds on false positive rate and false negative rate are considered equal.  As the average conditional entropy per token for the generated text increases, fewer tokens are required to achieve the same false negative and positive rates.

\subsection{Watermarking with random initialization}
\label{subsec:christWRandomChunk}

\cite{Christ2023} proposed to initiate the watermarked text with a chunk of tokens $R$ that is randomly sampled from language model $\mathbb{M}^{b}$. 
The empirical entropy for $R$ is defined as
\begin{align}
    H_e (\mathbb{M}_b, \PROMPT^b, R)
   =\sum_{i=1}^n -\ln p_i(w^b_{i}) .
    \label{eq:empEntropy}
\end{align}
After sampling tokens from language model $\mathbb{M}^{b}$ until  $H_e (\mathbb{M}_b, \PROMPT^b, R)$ exceeds a threshold $\lambda$ for the sampled set of tokens 
$R = w^b_{[m]}$, the watermarking encoding starts. 
The sampled binary tokens in $R$ and the context ngram $S_{i,h} = W^b_{[i-h:i-1]}$ are used as the input to a PRF 
$F_{\mathsf{sk}}: \{0,1\}^{\mathsf{poly}_1(\lambda)}\rightarrow\{0,1\}^{\mathsf{poly}_2(\lambda)}$, with a secret key $\mathsf{sk} \in \{0,1\}^\lambda$, to generate a random variable $Y_{i}$. 
Following the same steps as previous section, the watermarking encoding can be summarized using~\eqref{eq:watermarkingRuleChrist}, with $\by_{i} = F_{\mathsf{sk}}(R,S_{i,h})$.
Therefore, following this procedure for a prompt $\PROMPT^b$, a sequence of uniform random numbers, $\bY_{[n+1:\bL]}$ and a watermarked text $(R ,\bW^b_{[n+1:\mathbf{L}]})$ will be generated, where the initial chunk of tokens, $R$, is sampled from the language model $\mathbb{M}^{b}$ and the rest of the tokens are decided based on the watermarking encoding rule in~\eqref{eq:watermarkingRuleChrist} with $\by_{i} = F_{\mathsf{sk}}(R, S_{i,h})$. We then obtain the encoder in Algorithm~\ref{alg:ChristWinitial}.

\begin{algorithm}[t]
\caption{Zero-bit encoder with random initialization}
\label{alg:ChristWinitial}
\begin{algorithmic}[1]
 \REQUIRE A prompt $\alpha$ and a secret key $\mathsf{sk}$
 \ENSURE Watermarked text $W^b_{[L]}$
 \STATE $t \leftarrow 1$; $H \gets 0$;
 \WHILE{$\mathsf{done} \neq w^b_{t-1} $} 
    \STATE $p_t(1) \gets p_{\mathbb{M}^b}(1 \mid W^b_{[t-1]}, \PROMPT^b)$; $p_t(0) \gets 1 \! - \! p_t(1)$
    \IF{$H < \lambda$}
        \STATE Sample $w^b_t$ with $(p_t(0), p_t(1))$;
        \STATE $H \gets H - \ln{p_t(w^b_t)}$
        \IF{$H \geq \lambda~\mathsf{and}~ t\geq h$}
            \STATE $R \gets W^b_{[t]}$; $S_{t,h} \gets W^b_{[t-h,t-1]}$
        \ENDIF    
    \ELSE    
        \STATE Establish mapping rule $\Gamma_t(\Omega, \mathcal{V})$;\COMMENT{Eq.~\eqref{eq:christMappingRule}}
        \STATE $y_{t} \gets F_{\mathsf{sk}}(R,S_{t,h} )$;
            $w_t \gets \mathbbm{1}[y_{t} \in A_{2,t}]$;  
    \ENDIF   
    \STATE $t \gets t+1$;
\ENDWHILE
\end{algorithmic}
\end{algorithm}

The initial random chunk $R$, including its length, is random but does not depend on the output of the PRF. 
If the accumulated empirical entropy never surpasses $\lambda$, then the text generated by Algorithm~\ref{alg:ChristWinitial} will contain no watermark. 
Given a prompt $\PROMPT^b$ and a fixed initial chunk $R$, the conditional distribution for response of a language model $\mathbb{M}^b$, can be derived from~\eqref{eq:conditionDist}, with $\mathbb{M}$ and $\PROMPT$ replaced by $\mathbb{M}^b$ and $(\PROMPT^b, R)$, respectively. In other words, $(\PROMPT^b, R)$ is treated as the prompt (see Appendix~\ref{subsubsec:appendix_Wr_1}). 

As in the previous section, we can assume the decoder has access to the reconstructed sequence $Y_{[n+1:L]}$. For a given text $W^b$, a statistical test $\psi(W^b,Y)$ is used to test hypothesis $\mathcal{H}_0$ against $\mathcal{H}_1$. 
First, an initial chunk of $W^b$ with length $m$, is considered as $R = W^b_{[m]}$.
Then for that specific initial chunk $R$, $Y(R)=Y_{[m+1:L]}$ is constructed and we define \begin{align}
 C(W^b,Y;m) := \sum_{i=m+1}^{L}s(w^b_{i},y_{i}) .
 \label{eq:CWYr}
\end{align}
Again, in calculating $C(W^b,Y;m)$, any ngram ``context plus current token'' is considered only once.
The estimated $n$ in the decoder is defined as
\begin{align}
n^* :&= \argmin_{h\ \leq m \leq L-1}\left\{p\text{-value}_m(C(W^b,Y;m))\right\}
\nonumber \\
&=\min_{h\leq m \leq L-1}Q(L-m,C(W^b,Y;m)).
\label{eq:n*}
\end{align}
Then the global $p\text{-value}(C(W^b,Y))$, defined as,
\begin{align}
&p\text{-value}(C(W^b,Y)) \nonumber \\
&:= 1 - (1 - p\text{-value}_{n^*}(C(W^b,Y;n^*)))^{L-h},
\label{eq:globalP}
\end{align}
is calculated (see Appendix~\ref{subsubsec:appendix_Wr_2}).
If the global $p\text{-value}(C(W^b,Y))\leq \text{FPR}$, the text is detected as watermarked, where similar to the previous section, FPR is the maximum tolerable false positive rate. 
We then obtain the watermark detector given in Algorithm~\ref{alg:ChristWinitialDetector}.

\begin{algorithm}[t]
 \caption{Zero-bit detector with random initialization}
 \label{alg:ChristWinitialDetector}
 \begin{algorithmic}[1]
 \REQUIRE Text $W^b_{[L]}$ and a secret key $\mathsf{sk}$
 \ENSURE $\mathsf{true}$ or $\mathsf{false}$;
 \FOR {$m \gets h,\ldots, L-1$}
   \STATE $\mathcal{A}_m =\phi$; $C(W^b_{[L]}, Y; m) \gets 0$; $R \gets W^b_{[m]}$
   \FOR{$i\gets m+1,\ldots, L$}
      \IF{$W^b_{[i-h:i]} \notin \mathcal{A}_m$}
        \STATE Add $W^b_{[i-h:i]}$ to $\mathcal{A}_m$;   
            $y_{i} \gets F_{\mathbf{sk}}(R,W^b_{[i-1]})$
        \STATE $v_{i} \gets w^b_i\cdot y_{i} + (1 - w^b_i)\cdot (1 - y_{i})$
        \STATE $s(w^b_i,y_{i}) \gets \ln\frac{1}{v_i}$
       \STATE $C(W^b_{[L]}, Y; m) \gets C(W^b_{[L]}, Y; m) + s(w^b_i,y_{i})$
      \ENDIF 
  \ENDFOR
  \STATE $p_m \gets Q(|\mathcal{A}_m|, C(W^b_{[L]}, Y; m))$
\ENDFOR
\STATE $n^* \gets \argmin_{h \leq m \leq L-1} p_m$ \COMMENT{Eq.~\eqref{eq:n*}}
\IF{$1- (1-p_{n^*})^{L-h} \leq \text{FPR}$}
    \STATE \textbf{return} $\mathsf{true}$ 
           \textbf{Else return} $\mathsf{false}$
 \ENDIF
\end{algorithmic}
\end{algorithm}

As in the previous section, we can derive a lower bound $L_{\min}$ on the number of required watermarked tokens such that the false positive rate and false negative rate are bounded by FPR and FNR, respectively. 
At first we start with an estimate of $L_{\min}$ as $L_{\min} = L_0$, for a small value $L_0$. Then for this estimate $L_{\min} = L_0$, we derive $\beta$ and $\theta_n$ given in~\eqref{eq:pvalueCondontheta} and~\eqref{eq:WRthresholdLevel}. 
Using~\eqref{eq:falseNegativeRAteBoundr}, we derive an upper bound for false negative rate. 
If the derived upper bound on false negative rate is less than FNR, the estimated value of $L_{\min}$, is correct. 
However, if the derived upper bound on false negative rate is greater than FNR, we increase the estimate $L_{\min}$, i.e $L_{\min}  \gets L_{\min} + 1$, and repeat this process until for $L_{\min}=L^*$, the upper bound in~\eqref{eq:falseNegativeRAteBoundr} is bounded by FNR, and we derive $L_{\min} = \log{|\mathcal{V}|}\left\lceil\frac{L^*}{\log{|\mathcal{V}|}}\right\rceil$.

\section{Multi-Bit Distortion-free Watermarking}
\label{sec:watermarkWcoding}

In this section we develop a multi-bit distortion-free watermarking scheme based on the 
zero-bit schemes from Section~\ref{sec:christ.}. 
We first extend Definition~\ref{def:mappingRule} to a multi-bit watermarking mapping rule.

\begin{defn}[\textbf{Multi-bit watermarking mapping rule}]
For a random variable $\by$ defined over the sample space $\Omega$ with distribution $\Pr_{\by}$, a multi-bit watermarking mapping rule $\Gamma(\Omega, \mathcal{V}, \mathcal{M}; \Pr_{\by})$ maps a partition of a sample space $\Omega$ into the token set $\mathcal{V}$ of the language model $\mathbb{M}$. This process is done in two steps: First a partition $U(M)=\{A_1(M),\ldots, A_{|\mathcal{V}|}(M)\}$, depending on the message $M \in \mathcal{M}$, on the sample space $\Omega$ is formed. Then each partition part $A_j(M)$ is mapped to $v_j$ for $j=1,\ldots,|\mathcal{V}|$.
\label{def:codableMappingRule}
\end{defn}

The multi-bit watermarking mapping rule is depicted in Figure~\ref{fig:distotionFree}. The mapping rules in red and black depict two distinct mapping rules based on two different embedded messages. We assume $\mathcal{M}= \{0,1, \ldots, 2^m-1\}$, such that each message $M$ conveys $m$ bits of information. 
Let $\Gamma(\Omega, M)$ denote the partition on $\Omega$ depending the embedded message $M$, i.e. $\Gamma(\Omega, M) = U(M)$. 
On the other hand, $\Gamma(A_i(M))$, denoted the mapped token, i.e. $\Gamma(A_i(m)) = v_i$. 
Finally, $\Gamma(\Omega, \mathcal{V}, M)$ denotes the the process of partitioning the sample space based on the embedded message $M$ and then mapping the partition parts into the corresponding tokens (see Appendix~\ref{subsubsec:appendix_DISC_1}).

We can extend Proposition~\ref{prop:distortion-free} for a multi-bit watermarking mapping rule as follows.
\begin{prop}
A watermarking algorithm following a multi-bit watermarking mapping rule $\Gamma_t(\Omega, \mathcal{V}, \mathcal{M}; \Pr_{\by})$ is distortion-free if and only if for every 
prompt $\PROMPT$ and the past generated tokens $W_{[t-1]}$ and for every message 
$M \in \mathcal{M}$, 
\begin{align}
    \Pr\{\by_t \in A_{j,t}(M)\} &= p_{\mathbb{M}}(v_j \mid W_{[t-1]}, \PROMPT)
     = D_t[v_j] .
\end{align}
\label{prop:distortion-free&Codable}    
\end{prop}


Next, we propose a new multi-bit and distortion-free watermarking algorithm 
called Distribution Interval Shift Coding (DISC), which follows a multi-bit watermarking mapping rule as depicted in Figure~\ref{fig:DISCrule}. 
Given a prompt $\PROMPT$ and the past generated tokens $W^b_{[i-1]}$, the multi-bit watermarking mapping rule $\Gamma_i(\Omega, \mathcal{V}, \mathcal{M}; \Pr_{\by})$ is 
specified as follows:
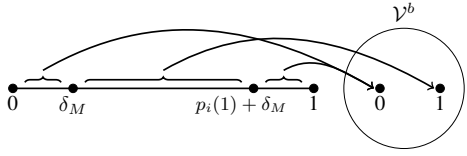
\begin{figure}[t]
    \centering
    \scalebox{0.8}{
    \begin{tikzpicture}
        \draw[thick] (-0,0) -- (5,0);
        \filldraw[black] (0,0) circle (2pt) node[anchor=north]{0};
        \filldraw[black] (5,0) circle (2pt) node[anchor=north]{1};
        \filldraw[black] (1,0) circle (2pt) node[anchor=north]{$\delta_M$};
        \filldraw[black] (4,0) circle (2pt) ;
        \node [below] at (3.8,0) {\small{$p_i(1) + \delta_M$}};
        \draw [->, thick] (0.5, 0.3) [out=30,in=150] to (6, 0);
        \draw [->, thick] (4.5, 0.3) [out=30,in=150] to (6, 0);
        \draw [->, thick] (2.5, 0.3) [out=30,in=150] to (7, 0);
        \filldraw[black] (6.1,0) circle (2pt) node[anchor=north]{0};
        \filldraw[black] (7.1,0) circle (2pt) node[anchor=north]{1};
        \draw (6.5, 0) circle (1cm);
        \node [above] at (6.5, 1) {$\mathcal{V}^b$};
        \draw [decorate,decoration = {brace}, thick] (0.2,0.1) --  (0.8,0.1);
        \draw [decorate,decoration = {brace}, thick] (4.2,0.1) --  (4.8,0.1);
        \draw [decorate,decoration = {brace}, thick] (1.2,0.1) --  (3.8,0.1);
    \end{tikzpicture}
    }
    \caption{Multi-bit watermarking mapping rule in DISC.} 
    \label{fig:DISCrule}
\end{figure}

\begin{align}
&\Omega = [0 , 1], \quad \quad \by_i\sim \Uniform[0,1],\nonumber \\
&\Gamma_i(\Omega,M) = \{A_{1,i}(M), A_{2,i}(M)\}, \nonumber \\
&\Gamma_i(A_{1,i}(M)) = 0, \quad \Gamma_i(A_{2,i}(M)) = 1 .
\label{eq:DISCMappingRuleP1}
\end{align}
For $p_i(1) + \delta_M \leq 1$,
\begin{align}
&A_{1,i}(M) =
    \{ 0  \leq \by_i < \delta_M\} \cup \{ p_i(1) + \delta_M \leq \by_i \leq 1\} , \nonumber \\
& A_{2,i}(M) = \{\delta_M \leq \by_i < p_i(1) + \delta_M\},
\end{align}
whereas for $p_i(1) + \delta_M > 1$, 
\begin{align}
&A_{1,i}(M) =
    \{  \delta_M + p_i(1) -1\leq \by_i < \delta_M \} , \nonumber \\
& A_{2,i}(M) = \{0 \leq \by_i < \delta_M + p_i(1) -1\}\cup\{\delta_M \leq \by_i < 1\} ,
\label{eq:DISCMappingRuleP2}
\end{align}
where $\delta_M = M\delta$,  
$M \in \mathcal{M} = \{0,1,\ldots, 2^m -1 \}$, and $\delta = 2^{-m}$. 
As in Section~\ref{sec:christ.}, $\by_{i} = F_{\mathsf{sk}}(R, S_{i,h}) \in [0,1]$ and the DISC 
scheme can be shown to be distortion-free.

For a prompt $\PROMPT^b$ and a message $M$, a sequence of uniform random numbers, $\bY_{[n+1:\bL]}$ and a watermarked 
text $\bW^b = ( R, \bW^b_{[n+1:\bL]})$ will be generated, where the initial chunk of tokens, $R$ 
is sampled from the language model $\mathbb{M}^{b}$, and the rest of the tokens are decided based on the watermarking encoding rule in~\eqref{eq:DISCMappingRuleP1} and~\eqref{eq:DISCMappingRuleP2} with $\by_i = F_{\mathsf{sk}}(R, S_{i,h})$. 
Therefore, we extend the encoding Algorithm~\ref{alg:ChristWinitial} in the $\mathbf{Input}$ and lines 13 and 15 to obtain the DISC encoder given in Algorithm~\ref{alg:DISCEncoder}.

\begin{algorithm}
 \caption{DISC encoder}
 \label{alg:DISCEncoder}
 \begin{algorithmic}
 \REQUIRE prompt $\alpha$, secret key $\mathsf{sk}$, message $M\in \mathcal{M}$
 \STATE \begin{small}13:\end{small} \quad Establish mapping rule $\Gamma_t(\Omega, \mathcal{V}, M)$;
 \STATE \begin{small}15:\end{small} \quad $w_t \gets \mathbbm{1}[y_t \in A_{2,t}(M)]$;
 \end{algorithmic}
\end{algorithm}

Similar to Section~\ref{subsec:christWRandomChunk}, for a given text $\{ W^b \}$, the statistical test $\psi(W^b,Y)$ is used to test hypothesis $\mathcal{H}_0$ against $\mathcal{H}_1$. 
This statistical test is performed as follows. 
First, an initial chunk of $W^b$ with length $m$, is considered as
$R = W^b_{[m]}$.
Then for that specific initial chunk $R$, $Y(R)= Y_{[m+1:L]}$ is constructed. Then for $\delta_{M'}= M'\delta$ and $M' \in \mathcal{M}$ as the assumed message by the decoder,
\begin{align}
 C(W^b, Y ;m, \delta_{M'}) = \sum_{i=m+1}^{L}s(w^b_{i},y_{i};\delta_{M'}),
 \label{eq:CWYrMultiBit}
\end{align}
is calculated. 
As in Section~\ref{sec:christ.}, any ngram ``context + current token'' is considered only once.
The score function is given as follows (see Figure~\ref{fig:DISC_score} in Appendix~\ref{subsubsec:appendix_DISC_2}):
\begin{align}  
s(w^b_{i}, y_i ; \delta_{M'})
=\left\{
\begin{array}{ll}
\ln\frac{1}{y_i-\delta_{M'}+1} & y_i \in [0, \delta_{M'}], w^b_{i} = 1,
\\
\ln\frac{1}{y_i-\delta_{M'}} & y_i \in (\delta_{M'}, 1], w^b_{i} = 1,
\\
\ln\frac{1}{\delta_{M'} - y_i} & y_i \in [0, \delta_{M'}), w^b_{i} =0,
\\
\ln\frac{1}{\delta_{M'} - y_i+1} & y_i \in [\delta_{M'}, 1], w^b_{i} =0.
\end{array}
\right.
\label{eq:scoreDISC}
\end{align}
Similar to Section~\ref{subsec:christWRandomChunk}, after calculating $C(W^b,Y;m, \delta_{M'})$, the estimated $n$ and $M$ in the decoder, i.e., $n^*$ and $M^*$, defined as,
\begin{align}
n^* , M^*
:&= \argmin_{\substack{h\ \leq m \leq L-1 \\ M' \in \mathcal{M}}}\left\{p\text{-value}_{m,M'}(C(W^b,Y;m, \delta_{M'}))\right\}.
\nonumber \\
&=\min_{\substack{h\ \leq m \leq L-1 \\ M' \in \mathcal{M}}}Q(L-m,C(W^b,Y;m, \delta_{M'})).
\label{eq:n*M*}
\end{align}
Then the global $p\text{-value}(C(W^b,Y))$, defined as,
\begin{align}
&p\text{-value}(C(W^b,Y)) \nonumber \\
&:= 1-\! (1-|\mathcal{M}|p\text{-value}_{n^*,M^*}(C(W^b,Y;n^*,M^*)))^{L-h},
\label{eq:globalPDISC}
\end{align}
is calculated (see Appendix~\ref{subsubsec:appendix_DISC_2}).
If the global $p\text{-value}(C(W^b,Y))\leq \text{FPR}$, the text is detected as watermarked.

Therefore, we extend the detecting Algorithm~\ref{alg:ChristWinitialDetector} in the $\mathbf{Output}$ and lines 4, 12-15, 17, 19, 20-21 to obtain the DISC decoder given in Algorithm~\ref{alg:DISCDecoder}.

\begin{algorithm}
 \caption{DISC decoder}
 \label{alg:DISCDecoder}
 \begin{algorithmic}
 \ENSURE $\mathsf{true}$ or $\mathsf{false}$ and message $M$ if text is watermarked
 \STATE \begin{small}4:\end{small} \quad $C(W^b, Y; m,M') \gets 0$;
 \STATE \begin{small}12:\end{small}\quad\textbf{for}\quad{$M'\in\mathcal{M}$ \quad\textbf{do}} 
 \STATE 
 \begin{small}13:\end{small}
 \quad\quad 
 Calculate $s(w^b_{i}, y_i ;\delta_{M'})$
 \COMMENT{Eq.~\eqref{eq:scoreDISC}}
 \STATE \begin{small}14:\end{small}
 \quad $
 C(W^b, Y; m,M') \gets C(W^b, Y; m,M') + s(w^b_i,y_{i};\delta_{M'})$;
 \STATE \begin{small}15:\end{small}\quad \textbf{EndFor}
 \STATE \begin{small}17:\end{small}\quad$p_{m,M'} \gets Q(|\mathcal{A}_m|, C(W^b, Y; m,M'))$;
 \STATE \begin{small}19:\end{small}\quad$n^*,M^* \gets \argmin_{\substack{h \leq m \leq L-1\\M'\in\mathcal{M}}} p_{m,M'}$ 
 \STATE \begin{small}20:\end{small}\quad\textbf{If}\quad{$1- (1- |\mathcal{M}|p_{n^*,M^*})^{L-h} \leq \text{FPR}$\quad\textbf{then}} 
 \STATE \begin{small}21:\end{small}\quad\textbf{return} $\mathsf{true}$ and $M'$ 
           \quad \textbf{Else return} $\mathsf{false}$ \textbf{Endif}
 \end{algorithmic}
\end{algorithm}

Similar to Section~\ref{subsec:christWRandomChunk}, for a watermarked text $\mathbf{W}_{\mathsf{W}}^b$ with length $L$ and initial chunk $R$, and the constructed $\mathbf{Y}(R)$, generated as response to prompt $\PROMPT^b$, we can derive a lower bound on the number of required watermarked tokens, $L_{\min}$, such that the false positive rate and false negative rate are bounded by FPR and FNR, respectively (see Appendix~\ref{subsubsec:appendix_DISC_3}). 

For a watermarked text $\bW_{\mathsf{W}}^b$ with length $L$ and initial chunk $R =W^b_{[n]}$, and the constructed $\bY(R)$, generated as response to prompt $\PROMPT^b$, we can avoid the exhaustive search over all $M'\in \mathcal{M}$ in~\eqref{eq:n*M*} to find $M^*$.
For $m=n$, $C(\bW^b,\bY;m, \Delta)$ follows the pattern given in Figure~\ref{fig:muDISC}, i.e., having maximum at $\Delta=0$ and two valleys on each side. 
Because we know the pattern of $C(\bW^b,\bY;m, \Delta)$, in order to find $M^*$, we can simply calculate $C(\bW^b,\bY;m, \delta_{M'})$ for a small set of equally spaced $\delta_{M'} \in \mathcal{M}_s$, with $|\mathcal{M}_s|\ll |\mathcal{M}|$. 
After calculating $C(\bW^b,\bY;m, \delta_{M'})$ at $\delta_{M'} \in \mathcal{M}_s$, because of the patterns of $C(\bW^b,\bY;m, \Delta)$
are known, we can obtain a rough estimate of $M^*$ and then by a finer search we can derive the exact $M^*$.  
Therefore, for a text $\bW_{\mathsf{W}}^b$ with length $L$, the complexity of decoding Algorithm~\ref{alg:DISCDecoder} can be reduced to $O(L^2)$, and does not depend on the number of possible embedded information bits. 

\section{Experiments}
\label{sec:expr}
\begin{figure}[t]
\centering
\includegraphics[scale = 0.55]{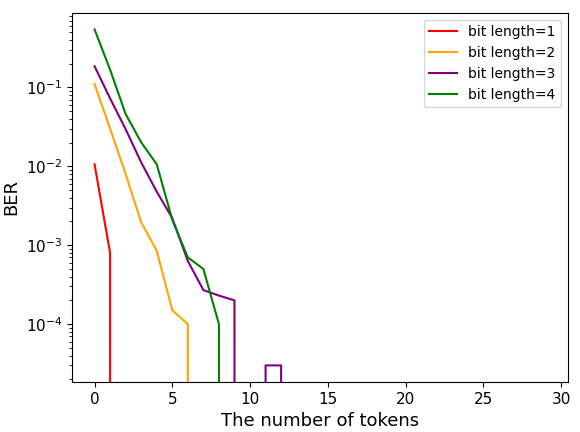}
\caption{BER when extracting bits with different length from the text with various length.}
\label{fig:BER}
\end{figure}

We assess the efficacy of DISC in embedding and extracting the watermark by simulating the binary sequences. Specifically, a real token is represented by 17~bits. For $m$-bits watermark, there exist $2^m$ distinct information options $M$ for watermarking, i.e., $M \in \mathcal{M} = \{0,1,\ldots, 2^m -1 \}$. With $M$ as the watermarking information to be conveyed, ${\delta}_{M}=M\delta$ should be embedded during text generation. In this context, we experiment with different values for $m$ ranging from 1 to 4.

For the text with $L$ bits (i.e., $L\mid17$~real tokens), we randomly generate the probability of each bit being 1 and a corresponding random value $u$ for that bit. Subsequently, the text is generated using the DISC encoder. In the process of watermark decoding, we investigate various ${\delta}_{M\prime}$ values to pinpoint the one exhibiting the highest score within the text.
Figure~\ref{fig:BER} shows the BER at a logarithmic scale when extracting watermarks of varying lengths across different numbers of real tokens in the text. For each length of text, the DISC algorithm was executed 10,000 times to compute BER. Notably, the BER for four different $m$ exhibits a significant decrease initially. Specifically, the extraction of a 1-bit watermark achieves a 0 BER over a text of merely 6 tokens, while a 4-bit watermark attains 0 BER at 20 tokens.

\section{Conclusion}
\label{sec:con}

We developed a new watermarking algorithm for Large Language Models (LLMs) with a primary focus on achieving distortion-free embedding of multiple information bits into the watermark. Details on the mathematical justification and analysis of the proposed watermarking algorithms are provided in the appendices. The key contribution lies in providing embedding power without compromising their original functionality or quality. To the best of our knowledge, this is the first work achieving multi-bit distortion-free watermarking with efficient
information decoding. This advancement opens up avenues for many
applications in content authentication and communication security. 

Future research can address the issue of further increasing the embedding capacity of LLM watermarking algorithms. In addition, watermarking algorithms that avoid
the overhead of binarization, e.g., \cite{Aaronson2023}, are preferable. In ongoing work, we are modifying our proposed DISC algorithms to work directly with the original token set of the LLM.  There is substantial scope for further exploration and improvement of LLM watermarking methods.




\clearpage
\bibliography{IEEEabrv,bibliography}
\bibliographystyle{icml2024}

\newpage
\appendix
\onecolumn
\section{Preliminaries}
\label{app:additionalDerivation}

\subsection{Entropy of a language model}
\label{subsec:entropy_appendix}
The set of all possible responses generated by $\mathbb{M}$ to a prompt $\PROMPT$ is denoted by $\mathcal{W}(\mathbb{M},\PROMPT)$.
Also, the set of all possible responses generated by $\mathbb{M}$ to a prompt $\PROMPT$ with length greater than or equal to $k$, is denoted by $\mathcal{W}^+_k(\mathbb{M},\PROMPT)$. 
Hence, we have

\begin{align}
\mathcal{W}(\mathbb{M},\PROMPT) = \mathcal{W}^+_1(\mathbb{M},\PROMPT),
\quad \mathcal{W}^+_k(\mathbb{M},\PROMPT) = \{W_{[L]} \in \mathcal{W}(\mathbb{M},\PROMPT): L \geq k\}.
\label{eq:responseSet}
\end{align}
On the other hand, $\Pr\{\mathbb{M}(\PROMPT)_{[L]} = W_{[L]}\}$ is the probability of all the generated responses of a language model $\mathbb{M}$ to a prompt $\PROMPT$ that starts with $W_{[L]}$.

With some abuse of notation, we use $D_t$ in this paper to represent the probability distribution of a language model. The assumption is that there is no ambiguity about the prompt $\PROMPT$ and the sequence of generated tokens so far, i.e., $W_{[t-1]}$. 
When we talk about altering a language model, we mean $D_t \rightarrow D'_t$, where the prompt $\PROMPT$ and the sequence generated tokens so far, i.e., $W_{[t-1]}$, are fixed. 
The entropy of the response of a language model $\mathbb{M}$ to a prompt 
$\PROMPT$ is defined as 
\begin{equation}
    H( \PROMPT ) := \Ex_{\mathbb{M}(\PROMPT)}
    \{- \ln\Pr\{\mathbb{M}(\PROMPT) \} \} 
    \label{eq:entropyAppendix}
\end{equation}
Using~\eqref{eq:conditionDist} and~\eqref{eq:responseSet}, we can simplify $H(\PROMPT)$ as,
\begin{align}
 &H(\PROMPT) = \sum_{ W_{[L]} \in \mathcal{W}(\mathbb{M},\PROMPT)} -\Pr\{\mathbb{M}(\PROMPT)= W_{[L]}\}\ln \Pr\{\mathbb{M}(\PROMPT)= W_{[L]}\}
    \nonumber \\
    &= \sum_{ W_{[L]} \in \mathcal{W}(\mathbb{M},\PROMPT)} -\Pr\{\mathbb{M}(\PROMPT)= W_{[L]}\}\ln \prod_{t=1}^L p_{\mathbb{M}}(w_t \mid W_{[t-1]}, \PROMPT)
    \nonumber \\
    &= \sum_{ W_{[L]} \in \mathcal{W}(\mathbb{M},\PROMPT)} \sum_{t=1}^L -\Pr\{\mathbb{M}(\PROMPT)= W_{[L]}\}\ln p_{\mathbb{M}}(w_t \mid W_{[t-1]}, \PROMPT)
    =\sum_{L=1}^\infty\sum_{W_{[L]} \in \mathcal{V}^L} - \Pr\{\mathbb{M}(\PROMPT)_{[L]} = W_{[L]}\} \ln p_{\mathbb{M}}(w_L \mid W_{L-1}, \PROMPT)
    \nonumber \\
    &=\sum_{L=1}^\infty\sum_{W_{[L-1]} \in \mathcal{V}^{L-1}}  \Pr\{\mathbb{M}(\PROMPT)_{[L-1]} = W_{[L-1]}\}
    \cdot\sum_{w_L \in \mathcal{V}}-p_{\mathbb{M}}(w_L \mid W_{[L-1]}, \PROMPT) \ln p_{\mathbb{M}}(\bw_L \mid W_{[L-1]}, \PROMPT)
    \nonumber \\
    &=\sum_{L=1}^{\infty} \sum_{W_{[L-1]} \in \mathcal{V}^{L-1}} \Pr\{\mathbb{M}(\PROMPT)_{[L-1]} = W_{[L-1]}\}
    \cdot H(p_{\mathbb{M}}(\bw_L \mid W_{[L-1]}, \PROMPT)) 
    = \sum_{L=1}^{\infty} \Ex\{H(p_{\mathbb{M}}(\bw_L \mid \bW_{[L-1]}, \PROMPT)) \},
\label{eq:entropySimplified}
\end{align}
where $H(p_{\mathbb{M}}(\bw_L \mid W_{[L-1]}, \PROMPT)) $ is defined as the conditional entropy of the $L$-th token, given already generated tokens $W_{[L-1]}$, of response of a language model $\mathbb{M}$ to a prompt $\PROMPT$.
Hence, $\Ex\{H(p_{\mathbb{M}}(\bw_L \mid \bW_{[L-1]}, \PROMPT))\}$, is the average conditional entropy of the $L$-th token, conditioned on the past tokens, of response of a language model $\mathbb{M}$ to a prompt $\PROMPT$ . 
Therefore, according to~\eqref{eq:entropySimplified}, entropy of response of a language model $\mathbb{M}$ to a prompt $\PROMPT$, is the summation of the average conditional entropy of the $L$-th token, conditioned on the past tokens, for $L = 1,2,\ldots$.  

We define $\zeta_L(\PROMPT)$ as the token-average of average conditional entropy, conditioned on the past tokens, for up to the $L$-th token, of response of a language model $\mathbb{M}$ to a prompt $\PROMPT$, for $L = 1,2,\ldots$.  In other words, 
\begin{align}
&\zeta_L(\PROMPT) = \frac{1}{L}\sum_{i=1}^L \Ex\{H(p_{\mathbb{M}}(\bw_L \mid \bW_{[L-1]}, \PROMPT)) \}.
\label{eq:zetaL}  
\end{align}
We define $\zeta(\PROMPT)= \lim_{L\rightarrow\infty}\zeta_L(\PROMPT)$, as the average conditional entropy, conditioned on the past tokens, per token of response of a language model $\mathbb{M}$ to a prompt $\PROMPT$. We assume for large values of $L$, $\zeta_L(\PROMPT) \approx \zeta(\PROMPT)$.

\subsection{Watermarking mapping rule}
\label{subsec:watermarking_mapping_appendix}

A watermarking mapping rule is shown in Figure~\ref{fig:distotionFree}.
In our notation, $\Gamma(\cdot)$ and $\Gamma(\cdot, \cdot)$ denote different operations. When it is used on a sample space, e.g. $\Gamma(\Omega)$, it denotes the partition formed on $\Omega$. 
In other words, $\Gamma(\Omega) = \{A_1,A_2,\ldots, A_{|\mathcal{V}|}\}$,
whereas when it is used on a partition part, e.g., $\Gamma(A_j)$, it denotes the mapped token, or $\Gamma(A_j) = v_j$. 
Finally, when it is used on a sample space and a token set, e.g., $\Gamma(\Omega, \mathcal{V})$, it means the whole process of partitioning the sample space and mapping partition parts to the tokens.

Note that the watermarking mapping rule is not stationary and throughout the text, the partitioning of the sample space and the mapping between partition parts to the different tokens changes. Hence, we denote the dependency of the mapping rule on the $t$-th token by adding a subscript $t$ as in $\Gamma_t(\cdot)$, $A_{j,t}$, and $\by_t$. However, we assume all $\by_t$'s are defined over the same sample space, and they have the same distribution.  
A watermarking encoding following a watermarking mapping rule is given in Algorithm~\ref{alg:distortion-freeAl}.

\begin{algorithm}
 \caption{Generating a watermarked text using a watermarking mapping rule $\Gamma(\Omega, \mathcal{V}; \Pr_{\by})$}
 \label{alg:distortion-freeAl}
 \begin{algorithmic}[1]
 \REQUIRE A prompt $\PROMPT$
 \ENSURE Watermarked text $W $
 \STATE $t \leftarrow 1$;
 \WHILE {$w_{t-1} \neq \mathsf{done} $}
    \STATE Establish mapping rule $\Gamma_t(\Omega, \mathcal{V})$;
    \STATE Generate $y_t\sim \Pr_{\by}$
    \FOR {$j \gets 1, \ldots, |\mathcal{V}|$} 
        \IF{{$y_t \in A_{j,t}$}}
            \STATE $w_t \gets v_j$;
            \STATE \textbf{Break}
        \ENDIF    
    \ENDFOR 
    \STATE $t \gets t+1$;
 \ENDWHILE
 \end{algorithmic}
\end{algorithm}

\section{Zero-bit Distortion-Free Watermarking}

\subsection{Binarization of language models}
\label{subsec:Binarization_appendix}

Let $E:\mathcal{V}\rightarrow\{0,1\}^{\log|\mathcal{V}|}$, denote an encoding function, which converts each token into a binary string. For example, $E(w)= (w^b_{1}, w^b_{2},\ldots, w^b_{\log|\mathcal{V}|})$.
Also, let $E(w)_n$ denote the $n$-th bit in the binary form of $w$.
Then, given a prompt $\PROMPT$ and the past generated tokens $W_{[t-1]}$, the distribution $D_t$ for language model $\mathbb{M}$, is converted into a series of probability distributions $D^b_t= (D^b_{t,1}, D^b_{t,2}, \ldots, D^b_{t,\log|\mathcal{V}|})$, where $D^b_{t,k} = (p_{t,k}(0), p_{t,k}(1))$, for $k=1, \ldots, \log|\mathcal{V}|$, for the language model $\mathbb{M}^{b}$. 
Here, 
\begin{align}
p_{t,k}(1) = \Pr\{\bw^b_{t,k} \! = \! 1 \mid W^b_{t,[k-1]}, W_{[t-1]}, \PROMPT \}, 
\quad \quad p_{t,k}(0) = 1 -p_{t,k}(1)  .
\label{eq:p_tk}
\end{align}
Note that, given a prompt $\PROMPT$ and the past generated tokens $W_{[t-1]}$, $p_{t,k}(0)$ can be simply calculated using $D_t$ and the encoding function $E(\cdot)$ as follows:
\begin{align}
p_{t,k}(1) &= \sum_{v \in \mathcal{V}} 
 \left \{ D_t[v] : E(v)_{[1:k]} \! = \!  (W^b_{t,[k-1]}, 1 ) 
 \right \} .
\label{eq:form_p_tk(1)}
\end{align}
In this way, sampling $\log|\mathcal{V}|$ times from the language model $\mathbb{M}^{b}$ is equivalent to sampling one token from language model $\mathbb{M}$.

Other coding schemes like \emph{Huffman Coding} can represent each token with fewer binary tokens on average.
However, they cannot be applied here, as these encoders require the probability distribution of each token, and as it was mentioned before, the underlying assumption in watermarking is that the decoder does not have access to the probability distributions.

Henceforth, we will use binary tokens to represent a text using the language model $\mathbb{M}^{b}$.
Hence, for convenience, we denote $p_{t,k}(1)$ and $w^b_{t,k}$ as $p_i(1)$ and $w^b_{i}$, respectively, where $i = (t-1)\log|\mathcal{V}| + k$.
We assume the prompt $\PROMPT$ and the past generated tokens $W_{[t-1]}$, are also represented by their $E(\cdot)$-encoded counterparts, i.e., $E(\PROMPT)$ and $E(W_{[t-1]})$, respectively. Henceforth, we also adopt the superscript $b$, to denote the binary equivalent of a prompt, sequence of tokens, etc, for example,  $E(\PROMPT) = \PROMPT^b$, $E(W_{[t-1]}) = W^b_{[t-1]}$. 

As binarization of the language model $\mathbb{M} \rightarrow \mathbb{M}^b$ using the encoding operator $E(\cdot)$ sets up a one-to-one correspondence
between $\mathbb{M}$ and $\mathbb{M}^b$, we have 
\begin{equation}
 H(\PROMPT)= \Ex_{\mathbb{M}^b (\PROMPT^b)} 
  \{-\ln\Pr\{\mathbb{M}^b(\PROMPT^b)\}\} .
    \label{eq:entropyBinary}   
\end{equation}
Using a similar approach as~\eqref{eq:entropySimplified}, we can further simplify~\eqref{eq:entropyBinary} as, 
\begin{align}
 H(\PROMPT)&
 = \sum_{i=1}^{\infty} \Ex\{H(p_{{\mathbb{M}^b}}(\bw^b_i \mid \bW^b_{[i-1]}, \PROMPT^b))\}
 =\sum_{i=1}^{\infty} \Ex\{H^b(\mathbf{p}_i(1)) \},   
 \label{eq:EntropyBinarySummation}
\end{align}
where $H^b(\mathbf{p}_i(1)) $ is defined as the conditional entropy of the $i$-th binary token, given already generated binary tokens $\bW^b_{[i-1]}$, of response of a language model $\mathbb{M}^b$ to a prompt $\PROMPT^b$ and is defined as
\begin{equation}
H^b(x) := -x \ln x  - (1-x)\ln(1-x).
\label{eq:conditionalBinaryEntropy}
\end{equation}
Here, $\mathbf{p}_i(1)$ is the probability of the $i$-th binary token being 1, given already generated tokens $\bW^b_{[i-1]}$, for response of a language model $\mathbb{M}^b$ to a prompt $\PROMPT^b$, and is defined in~\eqref{eq:p_tk} and~\eqref{eq:form_p_tk(1)}.
Additionally, $\Ex\{H^b(\mathbf{p}_i(1)) \}$, is the average conditional entropy of the $i$-th binary token, conditioned on the past tokens, of response of a language model $\mathbb{M}^b$ to a prompt $\PROMPT^b$. 
Similarly, $p_{\mathbb{M}^b}( W^b_{[0]} \mid \PROMPT^b ) = 1$, and $p_{\mathbb{M}^b}(\cdot \mid W_{[0]}^b, \PROMPT^b )= p_{\mathbb{M}^b}(\cdot \mid  \PROMPT^b )$.  
We can express $\Ex\{H(p_{\mathbb{M}}(\bw_L \mid \bW_{[L-1]}, \PROMPT))\}$ as summation of $\Ex\{H^b(\mathbf{p}_l(1)) \}$ for $l \in\{ (L-1)\log|\mathcal{V}|, \ldots, (L-1)\log|\mathcal{V}|+\log|\mathcal{V}|-1\}$ as follows,
\begin{align}
&\Ex\{H(\bw_L \mid \bW_{[L-1]}, \PROMPT))\} 
= \sum_{W_{[L-1]} \in \mathcal{V}^{L-1}}  \Pr\{\mathbb{M}(\PROMPT)_{[L-1]} = W_{[L-1]}\}
\cdot\sum_{w_L \in \mathcal{V}}-p_{\mathbb{M}}(w_L \mid W_{[L-1]}, \PROMPT) \ln p_{\mathbb{M}}(w_L \mid W_{[L-1]}, \PROMPT)
\nonumber \\
&= \sum_{W_{[L-1]} \in \mathcal{V}^{L-1}}  \Pr\{\mathbb{M}(\PROMPT)_{[L-1]} = W_{[L-1]}\} 
\cdot \sum_{j=1}^{\log|\mathcal{V}|}\sum_{W_{[i+1:i+j-1]}^b \in \mathcal{V}^{b^{j-1}}} p_{\mathbb{M}^b}(W^b_{[i+1:i+j-1]}\mid W_{[L-1]}^b, \PROMPT^b)
\nonumber \\
&\cdot \sum_{w^b_{i+j} \in \mathcal{V}^b}-p_{\mathbb{M}^b}(w^b_{i+j}\mid W^b_{[i+1:i+j-1]},W^b_{[i-1]},\PROMPT^b)
\cdot\ln p_{\mathbb{M}^b}(w^b_{i+j}\mid W^b_{[i+1:i+j-1]},W^b_{[i-1]},\PROMPT^b)
\nonumber \\
& = \sum_{j=1}^{\log|\mathcal{V}|} \Ex\{H^b(\mathbf{p}_{i+j}(1)) \}
,
\label{eq:EH&EHb}
\end{align}
where $i = (L-1)\log|\mathcal{V}|$.
Following the same steps as~\eqref{eq:zetaL}, we define $\zeta^b_L(\PROMPT^b)$ as the binary-token-average of average conditional entropy, conditioned on the past tokens, for up to the $L$-th binary token, of response of a language model $\mathbb{M}^b$ to a prompt $\PROMPT^b$, for $L = 1,\ldots$. Hence, 
\begin{align}
&\zeta^b_L(\PROMPT^b) = \frac{1}{L}\sum_{i=1}^L \Ex\{H^b(\mathbf{p}_{i}(1)) \}.
\label{eq:zetaiBinary}  
\end{align}
Similarly, we define $\zeta^b(\PROMPT^b)= \lim_{L\rightarrow\infty}\zeta^b_L(\PROMPT^b)$, as the average conditional entropy, conditioned on the past tokens, per binary token of response of a language model $\mathbb{M}^b$ to a prompt $\PROMPT^b$. Here, we also assume for large values of $L$, $\zeta^b_L(\PROMPT^b) \approx \zeta^b(\PROMPT^b)$. According to~\eqref{eq:zetaL},~\eqref{eq:EH&EHb} and~\eqref{eq:zetaiBinary},
\begin{equation}
\zeta^b(\PROMPT^b) = \frac{1}{\log|\mathcal{V}|} \zeta(\PROMPT) .
\label{eq:zeta&zetab}
\end{equation}

\subsection{Watermarking without random initialization}
\label{subsec:christWORandomChunk_appendix}

\subsubsection{part 1}
\label{subsubsec:appendix_WOr_1}

Assuming the watermarking process follows the described watermarking mapping rule and starts at the first token, then a watermarked-text 
${W}_{[L]}^b$ corresponds to a partition
part of an $L$-dimensional hypercube $[0,1]^L$.
For example, for a given prompt $\PROMPT^b$ and past generated tokens $W^b_{[i-1]}$, the event
$\{ \bw^b_i= 1 \}$ corresponds to the event $\{ \by_i\in [0,p_i(1))\} $. Hence, as this watermarking rule is distortion-free, we have
\begin{align}
\Ex \{H^b(\mathbf{p}_i(1)) \} = \Ex \{H^b(\mathbf{p}_i(1))\}, 
\label{eq:EH}
\end{align}
where $p_i(1)$ can be calculated using $W^b_{[i-1]}$ or $Y_{[i-1]}$. Note that in the LHS of~\eqref{eq:EH}, the expectation is calculated over the discrete distribution of $\bW^b_{[i-1]}$, whereas in the RHS, it is calculated over the continuous distribution of $\bY_{[i-1]}$. 
Henceforth the expectations are done with respect to $\bY$ and hence the subscript $\bY$ are omitted.
Hence,
\begin{equation}
 \zeta^b_L(\PROMPT^b) = \frac{1}{L}\sum_{i=1}^L \Ex\{H^b(\mathbf{p}_i(1))\}    .
 \label{eq:zetaBinaryY}
\end{equation}
Therefore, 
\begin{align}
H(\PROMPT) &=
\Ex_{\mathbb{M}^b(\PROMPT^b)}\{-\ln\Pr\{\mathbb{M}^b(\PROMPT^b)\}\}  
 = \sum_{i=1}^\infty \Ex\{H^b(\mathbf{p}_i(1))\} 
=\lim_{L\rightarrow\infty}L\zeta^b_L(\PROMPT^b) 
.
 \label{eq:entropyY}
\end{align}
Following this procedure for a prompt $\PROMPT$, a sequence of uniform random numbers, $\bY$ and a watermarked text $\bW^b$ will be generated. 
The general Algorithm~\ref{alg:distortion-freeAl} for a watermarking encoding following a watermarking mapping rule 
can then be updated to obtain Algorithm~\ref{alg:ChristWOinitial} for a Zero-bit distortion free watermarking encoding.

\begin{algorithm}[t]
 \caption{Zero-bit distortion-free watermarking encoder}
 \label{alg:ChristWOinitial}
 \begin{algorithmic}[1]
 \REQUIRE A prompt $\PROMPT$ and a secret key $\mathsf{sk}$
 \ENSURE Watermarked text $W^b$
 \STATE $t \leftarrow 1$;
 \WHILE {$\mathsf{done} \neq w^b_{t-1}$} 
    \STATE $p_t(1) \gets \Pr_{\mathbb{M}^b}(\bw^b_i=1 \mid w^b_{[t-1]}, \PROMPT^b)$;
    \STATE Establish mapping rule $\Gamma_t(\Omega, \mathcal{V})$;\COMMENT{Eq.~\eqref{eq:christMappingRule}}
    \STATE $y_t \gets F_{\mathsf{sk}}(S_{t,h})$;
    \STATE $w_t \gets \mathbbm{1}[y_t \in A_{2,t}]$;  
    \STATE $t \gets t+1$;
 \ENDWHILE
 \end{algorithmic}
\end{algorithm}

Suppose we have a watermarked text $W^b$ which was generated using the sequence of random numbers, $Y$, drawn from a uniform distribution. 
Assuming the secret key $\mathsf{sk}$ was shared with the decoder and no changes were made to the watermarked text, the decoder can reconstruct $Y$.
Note that for a fixed secret key $\mathsf{sk}$, the watermarking encoding in~\eqref{eq:watermarkingRuleChrist} always generates the same response for a prompt $\PROMPT$. This is an undesirable behaviour of a watermarking encoding and is addressed in the next section. 
Furthermore, the probabilities $\Pr\{C(\bW_{\mathsf{W}}^b, \bY) < \theta\}$ and $\Pr\{C(W_{\mathsf{NW}}^b, \bY) \geq \theta\}$ are calculated over $\bY$ and because watermarked text $\bW^b_{\mathsf{W}}$ depends on $\bY$, it is considered as random. However, the non-watermarked text, $W^b_{\mathsf{NW}}$, does not depend on $\bY$ and therefore is not considered random. In other words, all the derivations for non-watermarked texts hold for every fixed non-watermarked text. Although for a fixed $\PROMPT^b$ and a fixed $\mathsf{sk}$, $Y$ will be deterministic, we can assume $\bY$ is random due to uniform distribution over secret key $\mathsf{sk}$ or the use of true random function instead of a PRF.

\subsubsection{part 2}
\label{subsubsec:appendix_WOr_2}
\begin{figure}[t]
\centering
\includegraphics[scale = 0.55]{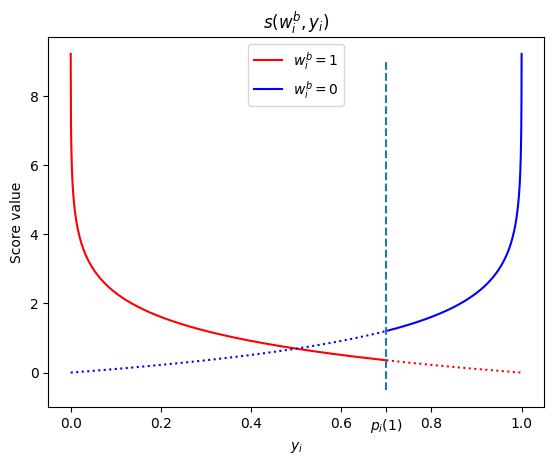}
\caption{Score function of the watermarking algorithm in~\cite{Christ2023}.}
\label{fig:codeless_score}
\end{figure}

\begin{figure}
\centering
\includegraphics[scale = 0.45]{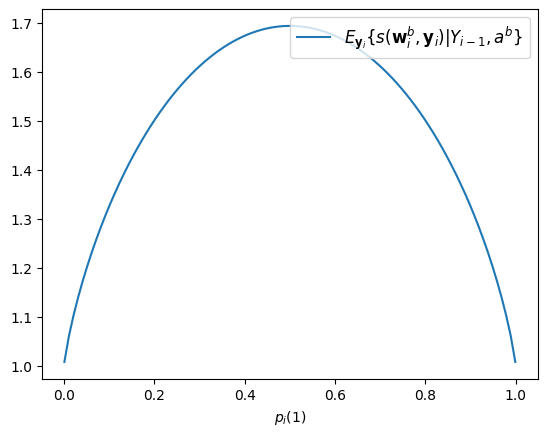}
\caption{Expectation and variance of the score value for a watermarked token.}
\label{fig:CodelessE}
\end{figure}

This score function is shown in  Figure~\ref{fig:codeless_score}. 
Given a prompt~$\PROMPT^b$, the expected value of the score function for the $i$-th token, if it is non-watermarked, can be derived as follows:
\begin{align}
\Ex  \{s(w^b_{i},\by_i) \mid \PROMPT^b\} 
&=\Ex \left\{\Ex  \left\{ s(w^b_{i},\by_i)\mid \bY_{[i-1]}, \PROMPT^b\right\}\right\} 
= \Ex\left\{ \int_0^{1} \ln\left(\frac{1}{v_i}\right) ~\mathrm{d}y_i \right\} = 1,
\label{eq:Es_nonwatermark}
\end{align}
where $v_i= w^b_i\cdot y_i + (1 - w^b_i)\cdot (1 - y_i)$.
On the other hand, given a prompt~$\PROMPT^b$, the expected value of the score function for the $i$-th token, if it is watermarked, can be derived as follows:
\begin{align}
\mu_i:=\Ex \{s(\bw^b_{i},\by_i)\mid \PROMPT^b \}  
=\Ex \left\{\Ex  \left\{ s(\bw^b_{i},\by_i)\mid \bY_{[i-1]}, \PROMPT^b\right\}\right\}. 
\label{eq:mus}
\end{align}
We can calculate $\Ex  \{ s(\bw^b_{i},\by_i)\mid \bY_{[i-1]}, \PROMPT^b\}$ as follows,
\begin{align}
 &\Ex  \left\{ s(\bw^b_{i},\by_i)\mid \bY_{[i-1]}, \PROMPT^b\right\} 
 = \int_0^{\mathbf{p}_i(1)} \ln\left(\frac{1}{y_i}\right) ~\mathrm{d}y_i 
 +  \int_{\mathbf{p}_i(1)}^{1} \ln\left(\frac{1}{1-y_i}\right)\! ~\mathrm{d}y_i  = 1 + H^b(\mathbf{p}_i(1)),
 \label{eq:Esyi}
\end{align}
where $H^b(\mathbf{p}_i(1))$ is defined in~\eqref{eq:conditionalBinaryEntropy}.
Therefore, we can simplify~\eqref{eq:mus} as
\begin{align}
    \mu_i = 1 + \Ex\left\{H^b(\mathbf{p}_i(1))\right\}.
    \label{eq:mussimplified}
\end{align}
The variance of the score value for the $i$-th token, if it is non-watermarked, given a prompt~$\PROMPT^b$, is
$\Var \{s(w^b_{i},\by_i) \mid \PROMPT^b\}=1$.
However, the variance of the score value for the $i$-th token, if it is watermarked, given a prompt~$\PROMPT^b$, can be derived as follows,
\begin{align}
  &\sigma^2_i :=\Var \{s(\bw^b_{i},\by_i)\mid \PROMPT^b\} 
  =\Ex \left\{\Ex \left\{ s^2(\bw^b_{i},\by_i)\mid \bY_{[i-1]}, \PROMPT^b\right\}\right\} - \mu^2_i.
  \label{eq:vars}
\end{align}
Similarly, we can calculate $\Ex \{ s^2(\bw^b_{i},\by_i)\mid \bY_{[i-1]}, \PROMPT^b\}$ as,
\begin{align}
\Ex  \left\{ s^2(\bw^b_{i},\by_i)\mid \bY_{[i-1]}, \PROMPT^b\right\} &=  G(\mathbf{p}_i(1))  
+ 2H^b(\mathbf{p}_i(1)) +2,  
\label{eq:Es2yi}
\end{align}
where 
\begin{align}
G(x) := x \ln^2 x  + (1-x)\ln^2(1-x), 
\label{eq:G}
\end{align} 
Similarly, $p_i(1)$ can be calculated using $W^b_{[i-1]}$ or $Y_{[i-1]}$.
As $H(p) \geq 1$, for all $p \in [0,1]$, $\mu^2_i > \mu_i$. 
Therefore, by using~\eqref{eq:G} we can bound $\sigma^2_i$ as
\begin{equation}
    \sigma^2_i < \max_{p_i(1)} \left\{G(p_i(1))  
+ H^b(p_i(1)) +1 \right\} = 2.2 .
\label{eq:boundonVarScore}
\end{equation}
The conditional expected value of the score function for a the $i$-th token, if it is watermarked, given a prompt~$\PROMPT^b$ and the past generated tokens $W^b_{[i-1]}$, with respect to $p_i(1)$, is shown in Figure~\ref{fig:CodelessE}. 
As we can see in Figure~\ref{fig:CodelessE}, the score value for the watermarked token is greater on average than the non-watermarked token.  
 According to~\eqref{eq:mussimplified} and~\eqref{eq:Es2yi}, if the distribution of $\mathbf{p}_i(1)$ is known, the expected value and variance of the score function for a the $i$-th token, if it is watermarked, can be derived. Although we do not have any knowledge about the distribution of $\mathbf{p}_i(1)$, if we assume $\mathbf{p}_i(1)\sim \mathsf{Uniform}[0,1]$, we can derive $\mu_i= 1.5$ and $\sigma^2_i = 1.25$.

Given a prompt~$\PROMPT^b$ and the past generated tokens $W^b_{[i-1]}$, without loss of generality, let us assume that $p_i(0) \geq p_i(1)$. Then for $x \geq -\ln p_i(1)$, the conditional cumulative distribution function (cdf) of $s(\bw^b_{i},\by_i)$ for the $i$-th token, if it is watermarked, given a prompt~$\PROMPT^b$ and the past generated tokens $W^b_{[i-1]}$, can be derived as follows:
\begin{align}
    F&_{s(\bw^b_{i},\by_i)}\left(x \mid W^b_{[i-1]},\PROMPT^b\right) 
    = \Pr\left\{s(\bw^b_{i},\by_i) \leq x\mid W^b_{[i-1]},\PROMPT^b\right \}
    \nonumber \\
    &= \Pr\left \{s(\bw^b_{i},\by_i) \leq x \mid \bw^b_{i} = 1,W^b_{[i-1]},\PROMPT^b\right \} 
    \cdot \Pr\left\{\bw^b_{i} = 1\mid W^b_{[i-1]},\PROMPT^b\right\} 
    \nonumber\\
    &+ \Pr \left \{s(\bw^b_{i},\by_i) \leq x \mid  \bw^b_{i} = 0,W^b_{[i-1]},\PROMPT^b\right \} 
    \cdot \Pr \left \{\bw^b_{i} = 0\mid W^b_{[i-1]},\PROMPT^b \right \}
    \nonumber\\
    &= \Pr \left \{\frac{1}{\ln \by_i} \leq x \mid  
      \by_i \! \in \! [0, p_i(1)), W^b_{[i-1]},\PROMPT^b \right \}
    \cdot \Pr \left \{ \by_i \! \in \! [0, p_i(1))\mid W^b_{[i-1]},\PROMPT^b \right \} \nonumber\\
    &= \Pr \left \{\frac{1}{\ln (1 \! - \! \by_i)} \leq x \mid 
    \by_i \! \in \!  [p_i(1),1] ,W^b_{[i-1]},\PROMPT^b\right \} 
    \cdot \Pr\left\{ \by_i \! \in \! [p_i(1),1]\mid W^b_{[i-1]},\PROMPT^b  \right\}
     = 1- 2e^{-x}.
    \label{eq:Fs_case1}
\end{align}

On the other hand, for $-\ln p_i(1) > x \geq -\ln p_i(0)$, we have 
\begin{align}
 F&_{s(\bw^b_{i},\by_i)}\left(x\mid W^b_{[i-1]},\PROMPT^b\right) 
 =\Pr \left \{s(\bw^b_{i},\by_i) \leq x\mid W^b_{[i-1]},\PROMPT^b\right \} 
 \nonumber\\
 &= \Pr \left \{s(\bw^b_{i},\by_i) \leq x \mid \bw^b_{i} = 1, W^b_{[i-1]},\PROMPT^b \right \} 
 \cdot \Pr \left \{\bw^b_{i} = 1\mid W^b_{[i-1]},\PROMPT^b \right \} \nonumber\\
 &+ \Pr \left \{s(\bw^b_{i},\by_i) \leq x \mid  \bw^b_{i} = 0, W^b_{[i-1]},\PROMPT^b\right \} 
 \cdot \Pr \left \{\bw^b_{i} = 0\mid W^b_{[i-1]},\PROMPT^b \right \} 
     =  p_i(0) - e^{-x} .
    \label{eq:Fs_case2}
\end{align}
Using \eqref{eq:Fs_case1} and~\eqref{eq:Fs_case2}, the conditional cumulative distribution function (cdf) of $s(\bw^b_{i},\by_i)$ for the $i$-th token, if it is watermarked, given a prompt~$\PROMPT^b$ and the past generated tokens $W^b_{[i-1]}$, is,
\begin{align}
 &F_{s(\bw^b_{i},\by_i)}\left(x\mid W^b_{[i-1]},\PROMPT^b\right) =\! \left\{\begin{array}{ll}
    \!\! p_i(0) - e^{-x}, & \! -\ln p_i(1)\! > x\! \geq\! -\ln p_i(0)\\
    \!\!  1- 2e^{-x},& \!x \geq -\ln p_i(1) .
 \end{array}\right.
 \label{eq:Fs}
\end{align}
Note that according to~\eqref{eq:Fs}, $s(\bw^b_{i},\by_i)$ is a continuous random variable. As we can see, $s(\bw^b_{i},\by_i)$ is not conditionally distributed as the summation of an exponential random variable and a constant, as was mentioned in~\cite{Christ2023}[Theorem 5].
On the other hand, given a prompt~$\PROMPT^b$ and the past generated tokens $W^b_{[i-1]}$, it can be easily shown $s(w^b_{i},\by_i)\sim \mathrm{Exp}(1)$, where $\mathrm{Exp}(\lambda)$ denotes the exponential distribution with parameter $\lambda$,
if it is non-watermarked,.

Therefore, using~\eqref{eq:mus} and~\eqref{eq:vars}, for the watermarked text $\bW_{\mathsf{W}}^b = \bW^b_{[L]}$ with a fixed length $L$, the constructed $\bY_{[L]}$, and a prompt~$\PROMPT^b$, we have,
\begin{align}
    &\vartheta:=\Ex\left\{C(\bW_{\mathsf{W}}^b, \bY) \mid \PROMPT^b\right\} 
    = \sum_{i=1}^{L} \Ex \{s(\bw_i^b,\by_i)\mid \PROMPT^b\}
    = \!L + \sum_{i=1}^{L}\Ex_{\bY_{i-1}}\left\{H^b(\mathbf{p}_i(1))\right\}
    = L + L\zeta^b_L(\PROMPT^b)
    \approx L + L\zeta^b(\PROMPT^b),
    \label{eq:Ec}
\end{align}
if $L$ is large enough. On the other hand, according to~\eqref{eq:boundonVarScore},
\begin{align}
\varsigma^2:=\Var\{C(\bW_{\mathsf{W}}^b&, \bY) \mid \PROMPT^b\}
=L\Var \{s(\bw^b_{i},\by_i)\mid \PROMPT^b\}  \leq 2.2L  .
\label{eq:boundonVarSumScore}
\end{align}

\subsubsection{part 3}
\label{subsubsec:appendix_WOr_3}

The CLT for the summation of a sequence of independent but not identically distributed random variables is referred to as the \textbf{Berry-Esse\'{e}n Theorem}~\cite{papoulis2002probability}[p. 283].

\begin{thm}[\textbf{Berry-Esse\'{e}n Theorem}]
    Let $\{ \mathbf{x}_i: i=1,2,\ldots \}$ be a sequence of independent random variables with bounded $\Ex\{\mathbf{x}_i\} = \mu_i$ and $
    \Var\{\mathbf{x}_i\} = \sigma^2_i$ and 
    \begin{equation}
        \Ex \{|\mathbf{x}_i-\mu_i|^3\}<c\sigma^2_i, \quad \forall i = 1, 2, \ldots ,
        \label{eq:CLTcondition}
    \end{equation}  
    where $c > 0$ is some constant. Then the cdf $F_{\mathbf{\bar{x}}}(x)$ of the normalized sum 
    \begin{equation}
        \bar{\mathbf{x}} = \frac{1}{\sigma}\sum_{i=1}^n (\mathbf{x}_i-\mu_i),
        \label{eq:normalizedSum}
    \end{equation}
    where $\sigma^2 = \sum_{i=1}^n \sigma^2_i$, converges to the cdf 
    $\Phi(x)$ of a zero-mean, unit variance
    normal distribution, i.e., $\mathcal{N}(0,1)$, with
    $\Phi(x) = \bigintsss_{-\infty}^x \frac{1}{\sqrt{2 \pi}} 
    e^{-\frac{u^2}{2}} ~\mathrm{d}u$. This convergence of the cdf is denoted as $\bar{\mathbf{x}} \xrightarrow{d}\mathcal{N}(0,1)$.
    \label{thm:CLT}
\end{thm}
According to~\eqref{eq:Fs}, given a prompt~$\PROMPT^b$ and the past generated tokens $W^b_{[i-1]}$, $s(\bw^b_{i},\by_i)$ for the $i$-th token, if it is watermarked, is defined for positive $x$ and has $e^{-x}$ terms. Therefore, for all $n=1,2,\ldots$,
\begin{align*}
 \Ex\left\{|s(\bw_i^b, \by_i)- \mu_i|^n\mid W^b_{i-1},\PROMPT^b\right\} < c_0, 
 \quad \Ex\left\{|s(\bw_i^b, \by_i)- \mu_i|^n\mid \PROMPT^b\right\} < c_1,
\end{align*}
for some $c_0,c_1>0$. Hence, the condition~\eqref{eq:CLTcondition} in Theorem~\ref{thm:CLT}, holds for some $c>0$. Therefore, according to~\eqref{eq:Ec}, for a watermarked text $\bW_{\mathsf{W}}^b$ and the constructed $\bY$
\begin{align}
\mathbf{C}(\bW_{\mathsf{W}}^b, \bY) \xrightarrow{d} \mathcal{N}\left(\vartheta, \varsigma^2\right),
\label{eq:sumScorePDFWatermarked}
\end{align}
where according to~\eqref{eq:boundonVarSumScore}, $\varsigma\leq 2.2L$. We can also use the CLT approximation for a non-watermarked text $W_{\mathsf{NW}}^b$ and the constructed $\bY$ as,
\begin{align}
\mathbf{C}(W_{\mathsf{NW}}^b, \bY) \xrightarrow{d} \mathcal{N}\left(L , L\right),
\label{eq:sumScorePDFNonWatermarked}
\end{align}
However, as mentioned before, for non-watermarked token $i$, $s(w^b_{i},\mathbf{y}_i)\sim \mathrm{Exp}(1)$. Hence, using the assumption of independence of $s(w^b_{i},\mathbf{y}_i)$'s, we can obtain a more accurate result as
\begin{align}
\mathbf{C}(W_{\mathsf{NW}}^b, \bY) \sim \mathbf{E}_L(1) .
\label{eq:sumScorePDFNonWatermarkedErlang}
\end{align}

\subsubsection{part 4}
\label{subsubsec:appendix_WOr_4}

For a given $L$ and FPR, the threshold in~\eqref{eq:thetaFPR}, can be computed using $\mathsf{gammainccinv}$ in SciPy~\cite{2020SciPy-NMeth}. The detection algorithm for this watermarking scheme is given in Algorithm~\ref{alg:ChristWOinitialDetector}, where returning $\mathsf{true}$ means the text was detected as watermarked, and returning $\mathsf{false}$ means the text was detected as non-watermarked.
\begin{algorithm}[t]
 \caption{Zero-bit distortion-free watermarking detector}
 \label{alg:ChristWOinitialDetector}
 \begin{algorithmic}[1]
 \REQUIRE A text $W^b_{[L]} $ ans a secret key $\mathsf{sk}$
 \ENSURE $\mathsf{true}$ or $\mathsf{false}$
 \STATE $\mathcal{A} =\phi$;
 \STATE $C(W^b_{[L]}, Y) \gets 0$;
 \FOR{$i\gets h+1,\ldots, L$}
    \IF{$S_{i+1,h+1}\notin \mathcal{A}$}
        \STATE Add $S_{i+1,h+1}$ to $\mathcal{A}$;    
        \STATE $y_i \gets F_{\mathsf{sk}}(S_{i,h})$;
        \STATE $v_i \gets w^b_i\cdot y_i + (1 - w^b_i)\cdot (1 - y_i)$;
        \STATE $s(w^b_i,y_i) \gets \ln\frac{1}{v_i}$
    \STATE $C(W^b_{[L]}, Y) \gets C(W^b_{[L]}, Y) + s(w^b_i,y_i)$;
    \ENDIF
\ENDFOR  
\STATE $\theta \gets Q^{-1}(|\mathcal{A}|,\text{FPR})$; \COMMENT{Eq.~\eqref{eq:thetaFPR}}
\IF{$C(W^b_{[L]}, Y) \geq \theta$} 
    \STATE \textbf{return} $\mathsf{true}$; 
\ELSE
    \STATE \textbf{return} $\mathsf{false}$; 
\ENDIF    
\end{algorithmic}
\end{algorithm}

For a given watermarked text $\bW^b_{\mathsf{W}}$ with length $L$, and the constructed $\bY$, using the normal approximation in~\eqref{eq:sumScorePDFWatermarked}, we can derive the false negative rate for the threshold given in~\eqref{eq:thetaFPR}, as 
\begin{align}
    \text{false negative rate} &= \Pr\{C(\bW_{\mathsf{W}}^b, \bY)<\theta \} \leq Q\left(\frac{\vartheta-\theta}{1.5\sqrt{L}}\right),
    \label{eq:falseNegativeRate}
\end{align}
where 
\begin{equation}
Q(x):=\frac{1}{\sqrt{2\pi}}\int_x^\infty e^{-\frac{u^2}{2}}~\mathrm{d}u.
\label{eq:Q}
\end{equation}

In order to derive a lower bound on the number of required watermarked tokens, $L_{\min}$, such that the false positive rate and false negative rate are bounded by FPR and FNR, respectively, we can use the following numerical method. 
At first we start with a estimate of $L_{\min}$ as $L_{\min} = L_0$, for a small value of $L_0$. For this estimate $L_{\min} = L_0$, we derive $\theta$ in~\eqref{eq:thetaFPR} using $\mathsf{gammainccinv}$. 
Then using the derived $\theta$, we derive an upper bound for false negative rate using~\eqref{eq:falseNegativeRate}.
If the derived upper bound is greater than FNR, we increase the estimate $L_{\min}$, i.e. $L_{\min} \gets  L_{\min} + 1$, and keep doing this process until for $L_{\min}=L^*$, the upper bound in~\eqref{eq:falseNegativeRate} is bounded by FNR, and we derive $L_{\min} = \log{|\mathcal{V}|}\left\lceil\frac{L^*}{\log{|\mathcal{V}|}}\right\rceil$.
 If $L_0$ is chosen small enough, the derived upper bound of the false negative for the corresponding $\theta$, will be always greater than FNR.
 Using the normal distribution approximation in~\eqref{eq:sumScorePDFNonWatermarked} and the inequality 
\begin{equation}
Q(x)\leq \frac{1}{2}e^{-\frac{x^2}{2}},
\label{eq:Qinequality}
\end{equation}
we derive an estimate of $\theta$ given by
\begin{equation}
    \theta_{\mathcal{N}} = L + \sqrt{2L\ln{\frac{1}{2\text{FPR}}}} = L + a_1 \sqrt{L} \approx \theta.
    \label{eq:thetaFPRN}
\end{equation}

\subsection{Watermarking with initial random initialization}
\label{subsec:christWRandomChunk_appendix}

\subsubsection{part 1}
\label{subsubsec:appendix_Wr_1}

By extending the definition in~\eqref{eq:entropyAppendix}, the entropy of response of a language model $\mathbb{M}^b$ to a prompt $\PROMPT^b$ that starts with $R$, is derived as
\begin{align}
    &H(\PROMPT^b, R) := \Ex_{\mathbb{M}^b(\PROMPT^b, R)}
    \left\{- \ln\Pr\left\{\mathbb{M}^b(\PROMPT^b, R) \right\} \right\} = \sum_{i=1}^{\infty} \Ex\left\{H(p_{{\mathbb{M}^b}}(\bw^b_i \mid \bW^b_{[n+1:i-1]}, \PROMPT^b,r))\right\}
    = \sum_{i=n+1}^{\infty} \Ex\left\{H^b(\mathbf{p}_i(1))\right \},
 \end{align}
where $H^b(\mathbf{p}_i(1))$ is the conditional entropy of the $i$-th binary token, given already generated tokens $W^b_{[n+1:i-1]}$, of response of a language model $\mathbb{M}^b$ to a prompt $\PROMPT^b$ that starts with $R$ and can be derived using~\eqref{eq:conditionalBinaryEntropy}. Here $\mathbf{p}_i(1)$ is the probability of the $i$-th binary token being 1, given already generated tokens $W^b_{[n+1:i]}$, for response of a language model $\mathbb{M}^b$ to a prompt $\PROMPT^b$ that starts with $R$.
Similarly, $\Ex\{H^b(\mathbf{p}_i(1)) \}$ is the average conditional entropy of the $i$-th binary token, conditioned on the past tokens, of response of a language model $\mathbb{M}^b$ to a prompt $\PROMPT^b$ that starts with $R$.

Similarly, by extending~\eqref{eq:zetaiBinary}, we define $\zeta^b_L(\PROMPT^b,R)$ as the binary-token-average of average conditional entropy, conditioned on the past tokens, for up to the $L$-th binary token, of response of a language model $\mathbb{M}^b$ to a prompt $\PROMPT^b$ that starts with $R$, for $L=n+1,n+2,\ldots$. Hence,
\begin{align}
\zeta^b_L(\PROMPT^b,R) 
= \frac{1}{L-n}\sum_{i=n+1}^L \Ex\left\{H^b(\mathbf{p}_i(1)) \right\}.
\label{eq:zetaiBinaryWinitial}  
\end{align}
Similarly, we define $\zeta^b(\PROMPT^b,R) =\lim_{L\rightarrow\infty}\zeta^b_L(\PROMPT^b,R)$, as the average conditional entropy, conditioned on the past tokens, per binary token of response of a language model $\mathbb{M}^b$ to a prompt $\PROMPT^b$ that starts with $R$. 
Additionally, we assume for large values of $L$, $\zeta^b_L(\PROMPT^b,R) \approx \zeta^b(\PROMPT^b,R)$. 
Similarly, as the watermarking rule in here is distortion-free, we have
\begin{align}
\Ex\left\{H^b(\mathbf{p}_i(1)) \right\}
= \Ex\left\{H^b\left(\mathbf{p}_i(1)\right)\right\}, 
\label{eq:EHWinitial}
\end{align}
where $p_i(1)$ can be calculated using $Y_{[n+1:i-1]}$ and $R$. Therefore, similar to~\eqref{eq:entropyY}, we have
\begin{align}
&H(\PROMPT^b,R) =
\Ex_{\mathbb{M}^b(\PROMPT^b,R)}\left\{-\ln\Pr\left\{\mathbb{M}^b(\PROMPT^b,R)\right\}\right\}  
 \!=\! \sum_{i=n+1}^\infty \Ex\left\{H\left(\mathbf{p}_i(1)\right)\right\} 
\!=\!\lim_{L\rightarrow\infty}(L-n)\zeta^b_L(\PROMPT^b,R) 
.
 \label{eq:entropyYr}
\end{align}

\subsubsection{part 2}
\label{subsubsec:appendix_Wr_2}
The global $p\text{-value}(C(W^b,Y))$, is defined as, the probability of observing a non-watermarked text as extreme as having minimum $p\text{-value}= p_{n^*}$, i.e.
\begin{align}
p\text{-value}(C(W^b,Y)) :&=\Pr\left\{\bigcup_{m=h}^{L-1}\left\{C(\bW^b,\bY;m)>p\text{-value}^{-1}_m(p_{n^*})\right\}\right\}.
\label{eq:globalPR}
\end{align}
is calculated. If the global $p\text{-value}(C(W^b,Y))\leq \text{FPR}$, the text is detected as watermarked, where similar to the previous section, FPR is the maximum tolerable false positive rate. 

This leads to a critical region $D_c = \bigcup_{m=h}^{L-1}\{C(\bW^b,\bY;m) \geq \theta_m\}$, where $\mathcal{H}_0$ is rejected, and a region of acceptance, $D^c_c = \bigcap_{m=h}^{L-1}\{C(\bW^b,\bY;m) < \theta_m\}$, where $\mathcal{H}_0$ is accepted. 
Note that, for a watermarked text $W^b_{\mathsf{W}}$, with length $L$ and initial chunk $R =W^b_{[n]}$, out of all possible choices of $m$, only for $m=n$, the generated $Y(r)$ in the decoder will be equal to $Y_{[n+1:L]}$ generated in the encoder.
Therefore, for a watermarked text $\bW^b_{\mathsf{W}}$, for $m \neq n$, there is no correlation between $\bY(r)$ and $\bW^b_{\mathsf{W}}$. 

Note that, similar to the previous section, for every $m$, the events $\{C(\bW_{\mathsf{W}}^b, \bY;m) < \theta_m\}$ and $\{C(W_{\mathsf{NW}}^b, \bY;m) \geq \theta_m\}$ are defined over the sample space for $\bY(r)$, and hence, a non-watermarked text, $W_{\mathsf{NW}}^b$ is independent of $\bY(r)$.
On the other hand, for the watermarked text $\bW^b_{\mathsf{W}}$ with length $L$, with $R =W^b_{[n]}$, the event $\{C(\bW_{\mathsf{W}}^b, \bY;n) < \theta_n\}$ is defined over the sample space $\bY(r)=\bY_{[n+1:L]}$, or equivalently over $\bW^b_{[n+1:L]}$. 
This means, in the case $R =W^b_{[n]}$, the event $\{C(\bW_{\mathsf{W}}^b, \bY;n) < \theta_m\}$ is defined for all the watermarked texts with length $L$ that start with $R =W^b_{[n]}$. 

As $\bY(r)$ is constructed based on the initial chunk of tokens, $R =W^b_{[m]}$, therefore, for different values of $m$, $C(\bW^b,\bY;m)$ are independent from each other. 
Note that, in calculating $\Pr\{C(\bW^b,\bY;m)> \theta_m\mid \mathcal{H}_0\}$, using a similar approach as previous section, we can derive an exact and an approximate distribution for $C(W^b_{\mathsf{NW}},\bY;m)$ as, 
\begin{align}
C(W^b_{\mathsf{NW}},\bY;m) \sim  \Er_{L-m}(1), 
\label{eq:C(W,Y;n)Exactpdf}
\quad C(W^b_{\mathsf{NW}},\bY;m)\xrightarrow{d} \mathcal{N}\left(L-m , L-m\right),
\end{align}
for $m=h,\ldots,L-1$.
As mentioned before, for the watermarked text $\bW^b_{\mathsf{W}}$, with length $L$, and with $R =W^b_{[n]}$, for $m\neq n$, constructed $\bY(r)$ is independent of $\bW^b_{\mathsf{W}}$. Therefore,
\begin{align}
C(W^b_{\mathsf{W}},\bY;m) \sim  \Er_{L-m}(1), 
\label{eq:C(W,Y;n)ExactpdfWatermarkedWrongm}
\quad C(W^b_{\mathsf{W}},\bY;m)\xrightarrow{d} \mathcal{N}\left(L-m , L-m\right),
\end{align}
for $m\neq n$. 
Therefore,
\begin{align}
p_{n^*} = \min_{h\leq m \leq L-1}Q(L-m,C(W^b,Y;m)).  
\label{eq:pn*}
\end{align} 
Hence, we can simplify~\eqref{eq:globalPR} as,
\begin{align*}
&p\text{-value}(C(W^b,Y)) =1- \prod_{m=h}^{L-1}\Pr\left\{C(\bW^b,\bY;m)\leq Q^{-1}(L-m,p_{n^*})\mid \mathcal{H}_0\right\} 
= 1- (1- p_{n^*})^{L-h}.
\end{align*}
In other words, for a text $W_{[L]}^b$, if for $p_{n^*}$ we have,
\begin{equation}
p_{n^*} \leq \beta:=1 -(1- \text{FPR})^{\frac{1}{L-h}},
    \label{eq:pvalueCondontheta}
\end{equation}
the text is detected as watermarked.
Therefore, the set of threshold levels $\{\theta_h,\ldots,\theta_{L-1}\}$ are derived as,
\begin{align}
 \theta_m =  Q^{-1}(L-m,\alpha).
 \label{eq:WRthresholdLevel}
\end{align}

Similar to the previous section, the expected value of the score function for the $i$-th token, if it is watermarked, given a prompt $\PROMPT^b$, with initial chunk $R = W^b_{[n]}$ is
\begin{align}
\mu_i:=\Ex \left\{s(\bw^b_{i},\mathbf{y}_{i})\mid \PROMPT^b,R \right\} 
=\Ex \left\{\Ex \left\{ s(\bw^b_{i},\mathbf{y}_{i})\mid \bY_{[n+1:i-1]}, \PROMPT^b,R\right\}\right\} 
= 1 +\Ex\left\{H^b\left(\mathbf{p}_i(1)\right)\right\}.
\label{eq:musr}
\end{align}
Similarly, the variance of the score value for the $i$-th token, if it is watermarked, given a prompt~$\PROMPT^b$, with initial chunk $R = W^b_{[n]}$ is
\begin{align}
  \sigma^2_i :=\Var \{s(\bw^b_{i},\mathbf{y}_{i})\mid \PROMPT^b,R\} 
  &=\Ex \left\{\Ex  \left\{ s^2(\bw^b_{i},\mathbf{y}_{i})\mid \bY_{[n+1:i-1]}, \PROMPT^b,R\right\}\right\} - \mu^2_i
  \nonumber \\
  &<   \max_{p_i(1)} \left\{G(p_i(1))  
+ H^b(p_i(1)) +1 \right\} = 2.2.
  \label{eq:varsr}
\end{align}
Similar to the previous section, using~\eqref{eq:musr} and~\eqref{eq:varsr}, for the watermarked text $\bW_{\mathsf{W}}^b$ with length $L$ and initial chunk $R = W^b_{[n]}$, and the constructed $\bY(r) = \bY_{[n+1:L]}$, generated as response to prompt~$\PROMPT^b$, we have,
\begin{align}
    \vartheta_n:&=\Ex\left\{C(\bW_{\mathsf{W}}^b, \bY;n) \mid \PROMPT^b\right\} 
    = \sum_{i=n+1}^{L} \Ex \{s(\bw_i^b,\mathbf{y}_{i})\mid \PROMPT^b,R\}
    = L -n+ \sum_{i=n+1}^{L}\Ex\left\{H^b\left(\mathbf{p}_i(1)\right)\right\} 
    \nonumber \\
    &= L-n + (L-n)\zeta^b_L(\PROMPT^b,R)
    \approx L-n  + (L-n)\zeta^b(\PROMPT^b,R),
    \label{eq:Ecr}
\end{align}
if $L$ is large enough. On the other hand, 
\begin{align}
\varsigma_n^2:=\Var\left\{C(\bW_{\mathsf{W}}^b, \bY;n) \mid \PROMPT^b\right\} 
=(L-n)\Var \{s(\mathbf{w}^b_{i},\mathbf{y}_{i})\mid \PROMPT^b,R\}  \leq 2.2(L-n)  
\label{eq:boundonVarSumScorer}
\end{align}
Similar to the previous section, by using CLT we get, 
\begin{align}
C(\bW_{\mathsf{W}}^b, \bY;n) \xrightarrow{d} \mathcal{N}\!\left(\vartheta_n, \varsigma_n^2\right).
\label{eq:sumScorePDFWatermarkedr}
\end{align}
For a watermarked text $\bW_{\mathsf{W}}^b$ with length $L$ and initial chunk $R = W^b_{[n]}$, and the constructed $\bY(r)$, we can derive the false negative rate, as 
\begin{align}
 \text{false negative rate} = \Pr\left\{\bigcap_{m=h}^{L-1}\left\{C(\bW^b,\bY;m) < \theta_m\mid \mathcal{H}_1\right\}\right\},
 \label{eq:falseNegativeRateThetan}
\end{align}
where the set of threshold levels $\{\theta_h,\ldots,\theta_{L-1}\}$ are given in~\eqref{eq:WRthresholdLevel}. Using~\eqref{eq:C(W,Y;n)ExactpdfWatermarkedWrongm} and~\eqref{eq:sumScorePDFWatermarkedr}, we can simplify~\eqref{eq:falseNegativeRateThetan} further as,
\begin{align}
 \text{false negative rate} = \prod_{\substack{m=h\\m\neq n}}^{L-1} (1-Q(L-m,\theta_m))\cdot Q\left(\frac{\vartheta_n-\theta_n}{\varsigma_n}\right)
 \leq (1-\text{FPR})^{\frac{L-h-1}{L-h}}\cdot Q\left(\frac{\vartheta_n-\theta_n}{1.5\sqrt{L-n}}\right).  
 \label{eq:falseNegativeRAteBoundr}
\end{align}

\begin{figure}[t]
\centering
\includegraphics[scale = 0.45]{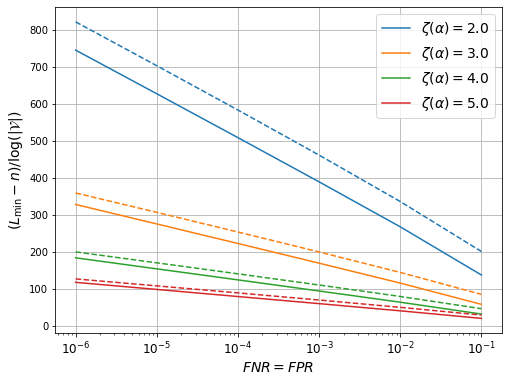}
\caption{Exact and approximate (dashed lines) $L_{\min}$ for $|\mathcal{V}|= 50272$, and $n = 3h\lceil\log|\mathcal{V}|\rceil$.}
\label{fig:LminR}
\end{figure}

As in Section~\ref{subsec:christWORandomChunk}, using the normal distribution approximation 
we derive $\theta_{\mathcal{N}}$, as an approximate of $\theta_n$ as,
\begin{equation}
    \theta_{\mathcal{N}} = L - n  + \sqrt{2(L-n)\ln{\frac{1}{2\beta}}}  \approx \theta_n.
    \label{eq:thetaFPRNWR}
\end{equation}
Similarly, using~\eqref{eq:thetaFPRNWR},~\eqref{eq:falseNegativeRAteBoundr}, we derive the approximation for $L_{\min}$ as,
\begin{align}
    \frac{f_1^2(\frac{\text{FPR}}{L_{\min}},\text{FNR})}{\zeta^b(\PROMPT^b)^2} \approx L_{\min} - n,
    \label{eq:LminApproxR}
\end{align}
where $f_1(\text{FPR},\text{FNR})$ is defined in~\eqref{eq:f1}. 
Therefore, according to~\eqref{eq:LminApproxR}, the initial random chunk $R$ has the equivalent effect of reducing FPR by a factor $L$, on the required number of watermarked tokens.
The exact and approximation value of $L_{\min}-n$, i.e. number of watermarked tokens, derived by the numerical method mentioned here and by using~\eqref{eq:LminApproxR}, for different values of $\zeta(\PROMPT^b)$ for vocabulary size $|\mathcal{V}|= 50272$, , and $|R|= 3h$, are shown in Figure~\ref{fig:LminR}, where the approximation values are depicted by dashed lines. In Figure~\ref{fig:LminR}, the bounds on false positive rate and false negative rate are considered equal. Similar to Section~\ref{subsec:christWORandomChunk}, as the average conditional entropy per token for the generated text increases, lower number of tokens are required to achieve the same false negative and positive rates. Comparing the results in Figures~\ref{fig:Lmin} and~\ref{fig:LminR} shows, having initial random chunk $R$, increases the required number of watermarked tokens by a factor about $1.4$. 

We can estimate $\zeta^b(\PROMPT^b,R)$ using already generated tokens in $R = W^b_{[n]}$ as,
\begin{align}
\zeta^b(\PROMPT^b,R) \approx   \frac{1}{n}\sum_{i=1}^{n} 
H(p_{{\mathbb{M}^b}}(w^b_i \mid \PROMPT^b,R_{[1:i-1]}))
=\frac{1}{n}\sum_{i=1}^{n}  H^b(p_i(1)) 
\label{eq:zetaEstimate}
\end{align}
where the RHS is the binary-token-average of conditional entropy, conditioned on the past tokens, for tokens in initial chunk $R$, of response of language model $\mathbb{M}^b$ for prompt $\PROMPT^b$. 

\section{Multi-bit Distortion-Free Watermarking}

\label{subsec:DISC_appendix}

\subsection{Part 1}
\label{subsubsec:appendix_DISC_1}

Here, we also denote the dependence of the multi-bit watermarking mapping rule on the $t$-th token by adding a subscript $t$. Then Algorithm~\ref{alg:distortion-freeAl} for a watermarking encoding following a watermarking mapping rule can be extended in the $\mathbf{Input}$ and lines 3 and 6 to obtain Algorithm~\ref{alg:distortion-free&CodableAl} for a watermarking encoding following a multi-bit watermarking mapping rule.

\begin{algorithm}
 \caption{Generating a watermarked text using a multi-bit watermarking mapping rule $\Gamma(\Omega, \mathcal{V}, \mathcal{M}; \Pr_{Y})$}
 \label{alg:distortion-free&CodableAl}
 \begin{algorithmic}
 \REQUIRE A prompt $\alpha$ and embedding message $M\in \mathcal{M}$
 \STATE \begin{small}3:\end{small} \quad\quad Establish mapping rule $\Gamma_t(\Omega, \mathcal{V}, M)$;
 \STATE \begin{small}6:\end{small} \quad\quad\quad\quad \textbf{If} $y_t \in A_{j,t}(M)$ \textbf{then}
 \end{algorithmic}
\end{algorithm}

\subsection{Part 2}
\label{subsubsec:appendix_DISC_2}

Define
\begin{equation}
    p_{n^*,M^*}:=p\text{-value}_{n^*,M^*}(C(W^b,Y;n^*, \delta_{M^*})).
    \label{eq:pn*M*}
\end{equation}
Then, the global $p\text{-value}(C(W^b,Y))$, is defined as
\begin{align}
p\text{-value}(C(W^b,Y))
 :
=\Pr\bigg\{\bigcup_{\substack{h \leq m \leq L-1\\M'\in \mathcal{M}}}&\bigg\{C(\bW^b,\bY;m,M')
> p\text{-value}^{-1}_{m,M'}(p_{n^*,M^*})\mid \mathcal{H}_0\bigg\}\bigg\}.
\label{eq:globalPDISCAppendix}
\end{align}
If the global $p\text{-value}(C(W^b,Y))\leq \text{FPR}$, the text is detected as watermarked. 
Similar to Section~\ref{subsec:christWRandomChunk}, for a watermarked text $\bW^b_{\mathsf{W}}$, with initial chunk $R =W^b_{[n]}$, there is no correlation between $\bY(r)$ and $\bW^b_{\mathsf{W}}$, for $m \neq n$, where $m$ is the length of the considered initial chunk in the decoder.  

\begin{figure}[t]
\centering
\includegraphics[scale = 0.55]{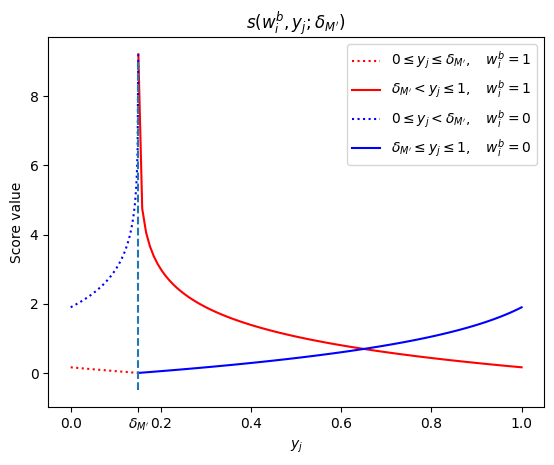}
\caption{Score function of the DISC algorithm.}
\label{fig:DISC_score}
\end{figure}
As in Section~\ref{subsec:christWRandomChunk}, we can see, for a non-watermarked text $W_{\mathsf{NW}}^b$ and for any initial chunk $R = W^b_{[m]}$, the constructed $\bY(r)$ is independent of $W_{\mathsf{NW}}^b$. 
Similarly, for different values of $m$, $C(\bW^b,\bY;m,\delta_{M'})$ are independent from each other. 

Therefore, we can simplify~\eqref{eq:globalPDISCAppendix} as
\begin{align}
p\text{-value}(C(W^b,Y))
:&= 1- \prod_{m=h}^{L-1}\bigcap_{M'\in \mathcal{M}}\Pr\Big\{C(\bW^b,\bY;m,\delta_{M'})
\leq  p\text{-value}^{-1}_{m,M'}(p_{n^*,M^*})\mid \mathcal{H}_0\Big\}
\nonumber \\
&= 1-
\prod_{m=h}^{L-1}\Pr\Big\{C_{\max}(\bW^b,\bY;m)\leq  p\text{-value}^{-1}_{m,M'}(p_{n^*,M^*})\mid \mathcal{H}_0\Big\},
\label{eq:PvalueGenDISCsimp}
\end{align}
where
\begin{equation}
C_{\max}(\mathbf{W}^b,\mathbf{Y},m) := \max_{M' \in \mathcal{M}}\left\{ \mathbf{C}(\mathbf{W}^b,\mathbf{Y};m,\delta_{M'})\right\}. 
\label{eq:Ctildemaxm}
\end{equation}
Note that, for a fixed $m$ and different values of $M'\in\mathcal{M}$, $C(\bW^b,\bY;m,\delta_{M'})$ are correlated with each other.
In fact, according to~\eqref{eq:scoreDISC}, for a fixed $m$ and all different values of $M'\in\mathcal{M}$, $s(\mathbf{w}^b_i,\mathbf{y}_{i};\delta_{M'})$ are functions of $s(\mathbf{w}^b_i,\mathbf{y}_{i};\delta_{0})$.
Hence, for a fixed $m$, by having $W^b$, $Y$ and $C(W^b,Y;m,\delta_{0})$ all $C(W^b,Y;m,\delta_{M'})$ can be uniquely determined for all $M' \in \mathcal{M}$.
Therefore, we cannot simplify the probability of intersection event in~\eqref{eq:PvalueGenDISCsimp} using product of probabilities. 

\begin{figure*}[t]
\centering
\subfigure[$m = L-1$, $w^b_{L}=1$, and $\theta_m = 3$.]{\label{fig:CDISC1D}        \includegraphics[scale = 0.55]{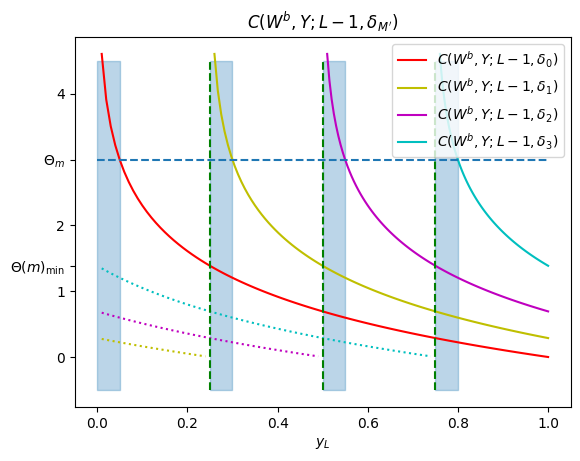}}
\subfigure[$m = L-2$, $(w^b_{L-1},w^b_{L})=(1,1)$, and $\theta_m = 5$.]{\label{fig:CDISC2D}\includegraphics[scale = 0.55]{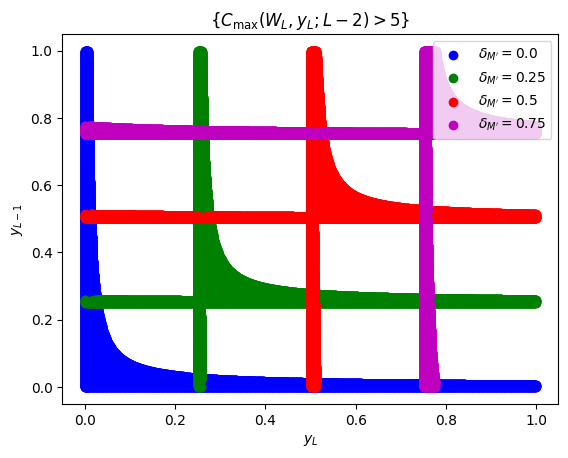}}
    \caption{$\left\{\mathbf{C}_{\max}(W_{\mathsf{NW}}^b,\mathbf{Y};m)> \theta_m\mid \mathcal{H}_0\right\}$ for $|\mathcal{M}|=4$.}
    \label{fig:CDISC}
\end{figure*}

The score function in~\eqref{eq:scoreDISC}, is just a shifted version of the score function in Section~\ref{sec:christ.}. Therefore, for a non-watermarked text $W^b_{\mathsf{NW}}$ with length $L$, $C(W^b_{\mathsf{NW}},\bY;m, \delta_{M'})$ follows the same distribution as $C(W^b_{\mathsf{NW}},\bY;m)$ in Section~\ref{subsec:christWRandomChunk}. Hence,
\begin{align}
C(W^b_{\mathsf{NW}},\bY;m, \delta_{M'}) \sim  E_{L-m}(1), 
\label{eq:C(W,Y;m,delta)Exactpdf}
\quad C(W^b_{\mathsf{NW}},\bY;m, \delta_{M'})\xrightarrow{d} \mathcal{N}\left(L-m , L-m\right),
\end{align}
for $m=h,\ldots,L-1$, and $M' \in \mathcal{M}$.
Similarly, for the watermarked text $\bW^b_{\mathsf{W}}$, with length $L$, and with $R =W^b_{[n]}$, for $m\neq n$ and $M' \in \mathcal{M}$,
\begin{align}
C(W^b_{\mathsf{W}},\bY;m,\delta_{M'}) \sim  E_{L-m}(1), 
\label{eq:C(W,Y;n,deltaExactpdfWatermarkedWrongm}
\quad C(W^b_{\mathsf{W}},\bY;m,\delta_{M'})\xrightarrow{d} \mathcal{N}\left(L-m , L-m\right).
\end{align}
We can simplify $p\text{-value}(C(W^b,Y))$ in~\eqref{eq:PvalueGenDISCsimp} using~\eqref{eq:C(W,Y;m,delta)Exactpdf}, as 
\begin{align}
\Pr\Big\{C_{\max}(\bW^b,\bY;m)> Q^{-1}(L-m,p_{n^*,M^*})\mid \mathcal{H}_0\Big\} 
&\approx |\mathcal{M}| \Pr\left\{C(W_{\mathsf{NW}}^b,\bY;m, \delta_M)> Q^{-1}(L-m,p_{n^*,M^*})\right\}
\nonumber \\
&=|\mathcal{M}|p_{n^*,M^*},   
\label{eq:PCmaxinequality}
\end{align}
for
\begin{align}
Q^{-1}(L-m,&p_{n^*,M^*}) \geq \theta(m)_{\min}:=(L-m)\ln\frac{1}{1-\delta_{|\mathcal{M}|-1}}
=-(L-m)\ln{\delta}= (L-m)r\ln{2}.  
\end{align}
The equality in~\eqref{eq:PCmaxinequality} stems from the symmetry of $C(W_{\mathsf{NW}}^b,\bY;m, \delta_{M'})$ for $M' \in \mathcal{M}$, over the sample space $\bY \in [0,1]^{L-m}$. 
This is shown in Figures~\ref{fig:CDISC1D} and~\ref{fig:CDISC2D} for $m = L-1$ and $m = L-2$, respectively, where the highlighted area shows the area for which $\{C_{\max}(W_{\mathsf{NW}}^b,\bY;L-1)> \theta\}$.
Therefore,
\begin{align}
p\text{-value}(C(W^b,Y)) \!=\!\left\lbrace\begin{array}{ll}\! 1-\! (1-|\mathcal{M}|p_{n^*,M^*})^{L-h}, &\!p_{n^*,M^*} \leq p_{\min}, \\
\!1 , &\!p_{n^*,M^*} > p_{\min},
\end{array}
\right.
\label{eq:pvalueDISC}
\end{align}
where 
\begin{align}
 p_{\min}&=\min_{h\leq m\leq L-1}Q(L-m,\theta(m)_{\min}) \approx Q(\sqrt{L-h}(r\ln{2}-1)),  
 \label{eq:pmin} 
\end{align}
where we have used the following lemma.
\begin{lemma}
For every $x>1$ and $m>n$, we have 
\begin{equation}
    Q(m,mx) < Q(n,nx). 
\end{equation}
\end{lemma}
In other words, for a text $W_{[L]}^b$, if for $p_{n^*,M^*}$ we have,
\begin{equation}
p_{n^*,M^*} \leq \beta_{\text{DISC}}:=\frac{1 -(1- \text{FPR})^{\frac{1}{L-h}}}{|\mathcal{M}|},
\label{eq:pvalueCondonthetaDISC}
\end{equation}
the text is detected as watermarked.
Therefore, similar to Section~\ref{subsec:christWRandomChunk}, the set of threshold levels $\{\theta_h,\ldots,\theta_{L-1}\}$ is
derived as follows:
\begin{align}
 \theta_m =  Q^{-1}(L-m,\alpha_{\text{DISC}}),~~m = h, \ldots, L-1,
 \label{eq:WRthresholdLevelDISC}
\end{align}
where the text is detected as non-watermarked if $\bigcap_{m=h}^{L-1}\{C_{\max}(W^b,Y;m) \leq \theta_m\}$, and it is detected as watermarked if, $\bigcup_{m=h}^{L-1}\{C_{\max}(W^b,Y;m) > \theta_m\}$. 

\subsection{Part 3}
\label{subsubsec:appendix_DISC_3}

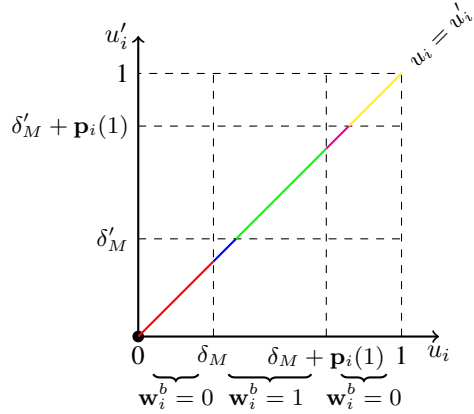
\begin{figure}[t]
    \centering
    \begin{tikzpicture}
    \draw[thick,<->] (4,0) -- (0,0) -- (0,4);
    \filldraw[black] (0,0) circle (2pt) node[anchor=north]{0};
    \node [left] at (0,4) {$u^{\prime}_{i}$};
    \node [below] at (4,0){$u_i$};
    \draw[dashed] (3.5,0) -- (3.5,3.5) -- (0,3.5);
    \node [below] at (3.5,0){1};
    \node [left] at (0,3.5){1};
    \draw[dashed] (1,0) -- (1,3.5);
    \node [below] at (1,0){\small{$\delta_M$}};
    \draw[dashed] (2.5,0) -- (2.5,3.5);
    \node [below] at (2.5,0){\small{$\delta_M +\mathbf{p}_i(1)$}};
    \draw[dashed] (0,1.3) -- (3.5,1.3);
    \node [left] at (0,1.3){\small{$\delta^{\prime}_M$}};
    \draw[dashed] (0,2.8) -- (3.5,2.8);
    \node [left] at (0,2.8){\small{$\delta^{\prime}_M + \mathbf{p}_i(1)$}};
    \draw [thick,red](0,0) -- (1, 1);
    \draw [thick,blue](1,1) -- (1.3, 1.3);
    \draw [thick,green](1.3,1.3) -- (2.5, 2.5);
    \draw [thick,magenta](2.5,2.5) -- (2.8, 2.8);
    \draw [thick,yellow](2.8,2.8) -- (3.5, 3.5);
    \node [rotate = 45] at (4,4){\small{$u_i = u^{\prime}_i$}};
    \draw [decorate,decoration = {brace, mirror}, thick] (0.2,-0.5) --  (0.8,-0.5);
    \node [below] at (0.5,-0.5) {\small{$\mathbf{w}^b_i = 0 $}};
    \draw [decorate,decoration = {brace, mirror}, thick] (1.2,-0.5) --  (2.3,-0.5);
    \node [below] at (1.7,-0.5) {\small{$\mathbf{w}^b_i = 1 $}};
    \draw [decorate,decoration = {brace, mirror}, thick] (2.7,-0.5) --  (3.3,-0.5);
    \node [below] at (3,-0.5) {\small{$\mathbf{w}^b_i = 0 $}};
\end{tikzpicture}
\caption{$u_i$ and $u'_i$ for DISC for $\delta_M + \mathbf{p}_i(1) \geq \delta^{\prime}_M \geq \delta_M$.} 
\label{fig:DISCEsCase1}
\end{figure}    

The expected value of the score function for the $i$-th token, if it is watermarked, given a prompt $\PROMPT^b$, with initial chunk $R = W^b_{[n]}$, and the embedded message $M$ in the encoder and the assumed message $M'$ in the decoder, is
\begin{align}
\mu_i:=\Ex \left\{s(\bw^b_{i},\by_{i}; \delta_{M'})\mid R,\PROMPT^b \right\} 
=\Ex\left\{\Ex  \left\{ s(\bw^b_{i},\by_{i}; \delta_{M'})\mid  R,\PROMPT^b,\bY_{[n+1:i-1]}\right\}\right\} .
\label{eq:musrDISC}
\end{align}
As mentioned before, $\by_i = F_{\mathsf{sk}}(R,S_{i,h})$ is generated in the encoder and then based on $\by_i$ and using~\eqref{eq:DISCMappingRuleP1} and~\eqref{eq:DISCMappingRuleP2}, $w^b_i$ is generated. 
In the decoder, on the other hand, $\by_i$ is generated again, and using~\eqref{eq:scoreDISC}, the score function $s(w^b_{i},y_{i};\delta_{M'})$ is calculated.
Assuming the secret key is shared between encoder and decoder and the watermarked text is not modified, the generated $\by_i$ in the encoder and the decoder are equal. 
This process is shown in Figure~\ref{fig:DISCEsCase1} for $\delta_M + \mathbf{p}_i(1) \geq \delta_{M'} \geq \delta_M$, where $\by_i$ generated in the encoder is denoted with $u_i$ and $\by_i$ generated in the decoder is denoted with $u'_i$.
Using Figure~\ref{fig:DISCEsCase1}, $\Ex  \{ s(\bw^b_{i},\by_{i}; \delta_{M'})\mid  R,\PROMPT^b,\bY_{[n+1:i-1]}\}$ is calculated as follows,
\begin{align}
&\Ex  \left\{ s(\bw^b_{i},\by_{i}; \delta_{M'})\mid \PROMPT^b,R,\bY_{[n+1:i-1]}\right\} =
 \int_0^{\delta_M} \ln\frac{1}{\delta^{\prime}_M - u} ~\mathrm{d}u
+\int_{\delta_M}^{\delta_{M'}} \ln\frac{1}{u - \delta_{M'} + 1} ~\mathrm{d}u 
 \nonumber \\
 &+\int_{\delta_{M'}}^{\delta_M + \mathbf{p}_i(1) } \ln\frac{1}{u - \delta_{M'}} ~\mathrm{d}u 
 + \int_{\delta_M + \mathbf{p}_i(1)}^{1} \ln\frac{1}{\delta_{M'} - u + 1} ~\mathrm{d}u
 = 1 - H^b(\Delta) + H^b(\mathbf{p}_i(1) - \Delta),
 \label{eq:EsDISCstep1}
\end{align}
where $H^b(x)$ is defined in~\eqref{eq:conditionalBinaryEntropy} and $\Delta = \delta_{M'}- \delta_M$.

It can be easily shown, for any continuous function $g(x)$, $\Ex  \{ g(s(\bw^b_{i},\by_{i}; \delta_{M'}))\mid  R,\PROMPT^b,\bY_{[n+1:i-1]}\}$ for the $i$-th token, if it is watermarked, given a prompt $\PROMPT^b$, with initial chunk $R=W^b_{[n]}$, and the embedded message $M$ in the encoder and the assumed message $M'$ in the decoder, is $\delta_M$-invariant, i.e.,
\begin{align}
 \Ex &\Big\{g(s(\bw^b_{i},\by_{i}; \delta_{M'}))\mid R,\PROMPT^b \Big\} 
 \!=\! \Ex \left\{g(s(\bw^b_{i},\by_{i}; \Delta))\mid R,\PROMPT^b \right\},
 \label{eq:deltaMinviariant}
\end{align}
with $\Delta = \delta_{M'} - \delta_{M}$. Using~\eqref{eq:deltaMinviariant}, we can calculate $\Ex  \{ s(\bw^b_{i},\by_{i}; \delta_{M'})\mid  R,\PROMPT^b,\bY_{[n+1:i-1]}\}$ for all the cases of $\delta_M$ and $\delta_M'$ as
\begin{align}
 \Ex  \left\{ s(\bw^b_{i},\by_{i}; \delta_{M'})\mid \PROMPT^b,R,\bY_{[n+1:i-1]}\right\}
 =\!\left\{\begin{array}{ll}
  \!1\! -\! H^b(\Delta)\! +\! H^b(\Delta - \mathbf{p}_i(1)), & \Delta \geq \mathbf{p}_i(1),
  \\
  \!1\! -\! H^b(\Delta)\! +\! H^b(\mathbf{p}_i(1) - \Delta),    & \mathbf{p}_i(1) \geq \Delta \!\geq 0,  
  \\
     \!1 \!-\! H^b(-\Delta)\! +\! H^b(\mathbf{p}_i(1)-\Delta ), & 0 \geq \Delta \geq -\mathbf{p}_i(0)
     \\
     \!1 \!-\! H^b(-\Delta)\! +\! H^b(-\mathbf{p}_i(0)-\Delta ),&  -\mathbf{p}_i(0) \geq \Delta.
 \end{array}\right.
 \label{eq:EsDISCDeltaCases}
\end{align}
\begin{figure}[t]
\centering
\includegraphics[scale = 0.55]{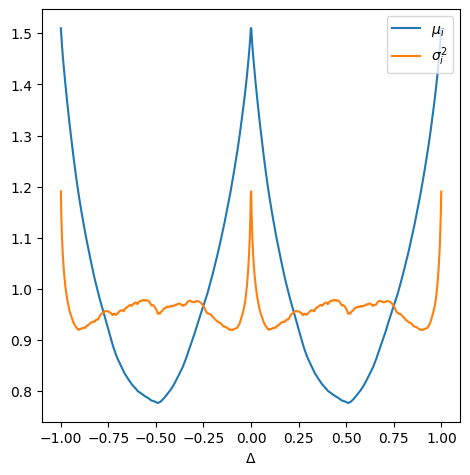}
\caption{$\mu_i$ and $\sigma^2_i$ for a watermarked token if $\mathbf{p}_i(1)\sim \mathsf{Uniform}[0,1]$.}
\label{fig:muDISC}
\end{figure}

Similarly, the variance of the score function for the $i$-th token, if it is watermarked, given a prompt~$\PROMPT^b$, with initial chunk $R=W^b_{[n]}$, and the embedded message $M$ in the encoder and the assumed message $M'$ in the decoder, for $\delta_M + \mathbf{p}_i(1) \geq \delta_{M'} \geq \delta_M$, is
\begin{align}
  \sigma^2_i :=\Var \{s(\bw^b_{i},\by_{i}; \delta_{M'})\mid \PROMPT^b,R\} 
  =  \Ex\left\{\Ex\left\{s^2(\bw^b_{i},\by_{i}; \delta_{M'})\mid \PROMPT^b,R , \bY_{[n+1:i-1]}\right\}\right\} 
  -\mu^2_i.
  \label{eq:varsrDISC}
\end{align}
Similarly, using~\eqref{eq:deltaMinviariant}, we can calculate $\Ex_{\by_{i}}  \{ s^2(\bw^b_{i},\by_{i}; \delta_{M'})\mid  R,\PROMPT^b,\bY_{[n+1:i-1]}\}$ for all the cases of $\delta_M$ and $\delta_M'$ as
\begin{align}
 \Ex  & \left\{ s^2(\bw^b_{i},\by_{i}; \delta_{M'})\mid \PROMPT^b,R,\bY_{[n+1:i-1]}\right\} \nonumber \\
 &=
 \left\{
 \begin{array}{ll}
2 - \mathbf{G}(\Delta) - 2H^b(\Delta) +  \mathbf{G}(\Delta-\mathbf{p}_i(1))  
 +2H^b(\Delta-\mathbf{p}_i(1)), & \Delta \geq \mathbf{p}_i(1),
\\
2 - \mathbf{G}(\Delta) - 2H^b(\Delta) +  \mathbf{G}(\mathbf{p}_i(1)-\Delta)
 + 2H^b(\mathbf{p}_i(1)-\Delta), & \mathbf{p}_i(1) \geq \Delta \geq 0,  
\\
2 - \mathbf{G}(-\Delta) - 2H^b(-\Delta) +  \mathbf{G}(\mathbf{p}_i(1)-\Delta) 
+ 2H^b(\mathbf{p}_i(1)-\Delta), 
& 
 0 \geq \Delta \geq -\mathbf{p}_i(0) ,
\\
2 - \mathbf{G}(-\Delta) - 2H^b(-\Delta) +  \mathbf{G}(-\mathbf{p}_i(0)-\Delta) 
+ 2H^b(-\mathbf{p}_i(0)-\Delta),
& -\mathbf{p}_i(0) \geq \Delta.
\end{array}
\right.
\label{eq:Es2DISCDeltaCases}
\end{align}

\begin{figure}[t]
\centering
\includegraphics[scale = 0.55]{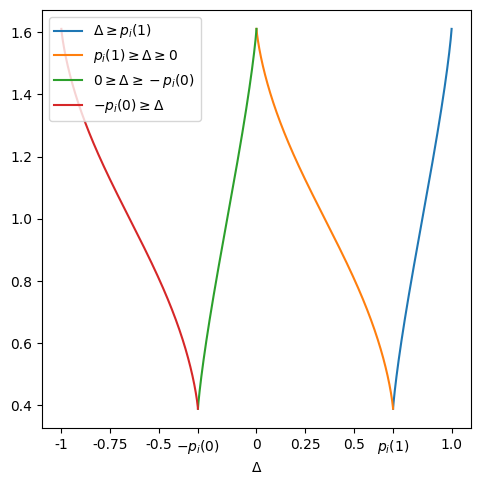}
\caption{$\Ex  \{ s(\bw^b_{i},\by_{i}; \delta_{M'})\mid  R,\PROMPT^b,Y_{[n+1:i-1]}\}$ with $p_i(1)=0.7$.}
\label{fig:EsDISCDelta}
\end{figure}
$\Ex  \{ s(\bw^b_{i},\by_{i}; \delta_{M'})\mid  R,\PROMPT^b,Y_{[n+1:i-1]}\}$ for a watermarked token, given a prompt $\PROMPT^b$, with initial chunk $R =W^b_{[n]}$, with $p_i(1) = 0.7$, and with embedded message $M$ in the encoder and the assumed message $M'$ in the decoder is shown in Figure~\ref{fig:EsDISCDelta}.

Similar to Section~\ref{subsec:christWORandomChunk}, if the distribution of $\mathbf{p}_i(1)$ is known, the expected value and variance of the score function for a the $i$-th token, if it is watermarked, can be derived. Similarly, if we assume $\mathbf{p}_i(1)\sim \mathsf{Uniform}[0,1]$, we can derive $\mu_i$ and $\sigma^2_i$, as shown in Figure~\ref{fig:muDISC}.
Note that for $\Delta=0$, $s(\bw^b_{i},\by_{i}; \delta_{M'})$ is equal to $s(\bw^b_{i},\by_{i})$ in Section~\ref{sec:christ.}. Hence, for $\Delta=0$, $\sigma^2_i  \leq 2.2$.
Therefore, for the watermarked text $\bW_{\mathsf{W}}^b$ with length $L$ and initial chunk $R =W^b_{[n]}$, and the constructed $\bY(r) =  \bY_{[n+1:L]}$, generated as response to prompt~$\PROMPT^b$, we have,
\begin{align}
   \vartheta_n:&= \Ex\left\{C(\bW_{\mathsf{W}}^b, \bY;n, \delta_M) \mid \PROMPT^b\right\} 
    = \sum_{i=n+1}^{L} \Ex_{\bY_{[n+1:i]}} \{s(\bw_i^b,\by_{i-n};\delta_M)\mid R,\PROMPT^b\}
    \nonumber \\
    &= L-n + (L-n)\zeta^b_L([\PROMPT^b,R])
    \approx L-n  + (L-n)\zeta^b([\PROMPT^b,R]),
    \label{eq:EcrDISC}
\end{align}
if $L$ is large enough. On the other hand, 
\begin{align}
\varsigma_n:=\Var\left\{C(\bW_{\mathsf{W}}^b, \bY;n,\delta_M) \mid \PROMPT^b\right\} :=
=(L-n)\Var \{s(\bw^b_{i},\by_{i};\delta_M)\mid \PROMPT^b,R\}  \leq 2.2(L-n)  
\label{eq:boundonVarSumScorerDISC}
\end{align}
Similarly, by using CLT we get, 
\begin{align}
C(\bW_{\mathsf{W}}^b, \bY;n, 
\delta_M) \xrightarrow{d} \mathcal{N}\!\left(\vartheta_n, \varsigma_n\right).
\label{eq:sumScorePDFWatermarkedrDISC}
\end{align}

\begin{figure*}[t]
\centering
\subfigure[$m = L-1$, $p_L(1)=0.7$, and $\theta_m(Z^*) = 3$.]{\label{fig:CH1DISC1D} \includegraphics[scale = 0.55]{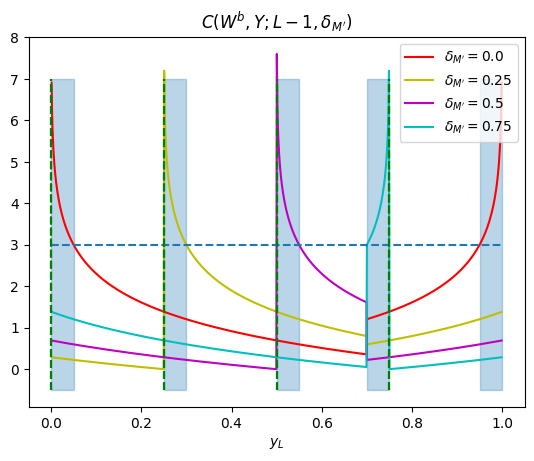}}
\subfigure[$m = L-2$, $(p_{L-1}(i),p_{L}(1))=(0.7,0.4)$, and $\theta_m = 5$.]{\label{fig:CH1DISC2D}\includegraphics[scale = 0.55]{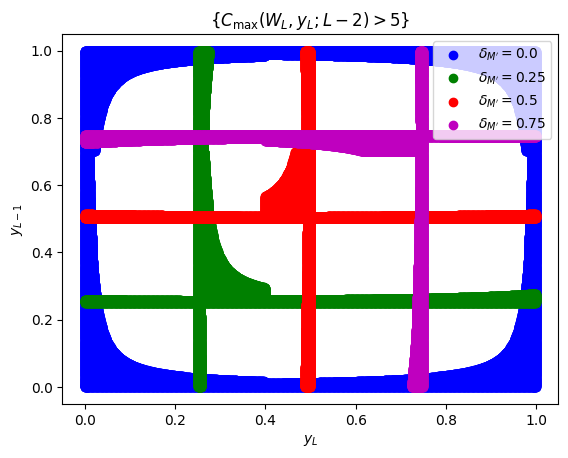}}
\caption{$\left\{\mathbf{C}_{\max}(W_{\mathsf{NW}}^b,\mathbf{Y};m)> \theta_m\mid \mathcal{H}_1\right\}$ for $|\mathcal{M}|=4$ and $\delta_M=0$.}
\label{fig:CH1DISC}
\end{figure*}

Similar to Section~\ref{subsec:christWRandomChunk}, for a watermarked text $\bW_{\mathsf{W}}^b$ with length $L$ and initial chunk $R =W^b_{[n]}$, and the constructed $\bY(r)$, with embedded message $M$ in the encoder and the assumed message $M'$ in the decoder,  false negative rate is 
\begin{align}
 \text{false negative rate} =\Pr\left\{\bigcap_{m=h}^{L-1}\left\{C_{\max}(\bW^b,\bY;m) < \theta_m\mid \mathcal{H}_1\right\}\right\},
 \label{eq:falseNegativeRateDISC}
\end{align}
where the set of threshold levels $\{\theta_h,\ldots,\theta_{L-1}\}$ are given in~\eqref{eq:WRthresholdLevelDISC}.  
The events $\{C_{\max}(W_{\mathsf{W}}^b,\bY;L-1)> \theta_m\}$ for $m = L-1$ and $m = L-2$ are shown in Figures~\ref{fig:CH1DISC1D} and~\ref{fig:CH1DISC2D}, respectively. 
As we can see, in Figure~\ref{fig:CH1DISC}, for the case $\Delta = 0$, all the corners of the $(L-m)$-dimensional cube $\bY \in[0,1]^{L-m}$, and the strips along each edge of this cube satisfy $\{C_{\max}(W_{\mathsf{W}}^b,\bY;L-1, \delta_0)> \theta_m\}$. 
These areas for $m= L-2$ are shown with blue shaded area in Figure~\ref{fig:CH1DISC2D}.
For other cases of $\Delta \neq 0$, however, just a volume of $\bY$ equal to $\delta_{M'}$-shifted of one of the corners of the $(L-m)$-dimensional cube and  the corresponding edges strips satisfies $\{C_{\max}(W_{\mathsf{W}}^b,\bY;L-1, \delta_0)> \theta_m(Z^*)\}$. 
For each case $\Delta = \delta_{M'}\neq 0$, these areas for $m= L-2$ are shown with shaded areas in green, red, or magenta in Figure~\ref{fig:CH1DISC2D}.
Therefore, using~\eqref{eq:sumScorePDFWatermarkedrDISC}, we can simplify~\eqref{eq:falseNegativeRateDISC}, as 
\begin{align}
\Pr\Big\{C_{\max}(\bW^b,\bY;n) > \theta\mid \mathcal{H}_1\Big\}   
&\approx \left((|\mathcal{M}|-1)\frac{L-n}{2^{L-n}}+1\right) 
\cdot\Pr\left\{C(\bW^b,\bY;n,\delta_{0}) > \theta_n\mid \mathcal{H}_1\right\}
\nonumber \\
&
=\left((|\mathcal{M}|-1)\frac{L-n}{2^{L-n}}+1\right)Q\left(\frac{\theta_n-\vartheta_n}{\varsigma_n}\right)
.   
\label{eq:PCmaxinequalityH1}
\end{align}
Therefore, using~\eqref{eq:C(W,Y;n,deltaExactpdfWatermarkedWrongm} and ~\eqref{eq:PCmaxinequalityH1} we can simplify
the false negative rate as
\begin{align}
&\text{false negative rate} = \prod_{\substack{m=h\\m\neq n}}^{L-1} (1-|\mathcal{M}|Q(L-m,\theta_m))
 \cdot \left(1-\left((|\mathcal{M}|-1)\frac{L-n}{2^{L-n}}+1\right)Q\left(\frac{\theta_n-\vartheta_n}{\varsigma_n}\right)\right)
 \nonumber \\
 &\leq (1-\text{FPR})^{\frac{L-h-1}{L-h}} 
 \cdot \left(1-\left((|\mathcal{M}|-1)\frac{L-n}{2^{L-n}}+1\right)Q\left(\frac{\theta_n-\vartheta_n}{1.5\sqrt{L-n}}\right)\right),  
 \label{eq:falseNegativeRAteBoundDISC}
\end{align}
with $\theta_n$ given in~\eqref{eq:WRthresholdLevelDISC}. Similar Section~\ref{subsec:christWRandomChunk}, for a watermarked text $\bW_{\mathsf{W}}^b$ with length $L$ and initial chunk $R =W^b_{[n]}$, and the constructed $\bY(R)$, generated as response to prompt $\PROMPT^b$, using~\eqref{eq:pvalueCondonthetaDISC} and~\eqref{eq:falseNegativeRAteBoundDISC}, we can derive a lower bound on the number of required watermarked tokens, $L_{\min}$, such that the false positive rate and false negative rate are bounded by FPR and FNR, respectively. 

As in Section~\ref{subsec:christWRandomChunk_appendix}, using the normal distribution approximation 
we derive $\theta_{\mathcal{N}}$, as an approximate of $\theta_n$ as,
\begin{equation}
    \theta_{\mathcal{N}} = L - n  + \sqrt{2(L-n)\ln{\frac{1}{2\beta_{\text{DISC}}}}}  \approx \theta_n.
    \label{eq:thetaFPRNDISC}
\end{equation}
Similarly, using~\eqref{eq:thetaFPRNDISC},~\eqref{eq:falseNegativeRAteBoundDISC}, we derive the approximation for $L_{\min}$ as,
\begin{align}
    \frac{f_1^2(\frac{\text{FPR}}{|\mathcal{M}|L_{\min}},\text{FNR})}{\zeta^b(\PROMPT^b)^2} \approx L_{\min} - n,
    \label{eq:LminApproxDISC}
\end{align}
where $f_1(\text{FPR},\text{FNR})$ is defined in~\eqref{eq:f1}. 
Therefore, according to~\eqref{eq:LminApproxDISC}, the effect of having initial random chunk $R$ and the embedding message set $\mathcal{M}$ in DISC, on the required number of watermarked tokens is equivalent to reducing FPR by a factor $L|\mathcal{M}|$.
The exact and approximation value of $L_{\min}-n$, i.e. number of watermarked tokens, derived by the numerical method mentioned here and by using~\eqref{eq:LminApproxDISC}, for different values of $\zeta(\PROMPT^b)$ for vocabulary size $\mathcal{V}= 50272$, , $|R| = 3h\lceil\log|\mathcal{V}|\rceil$, and $|\mathcal{M}|=2^{10}$, are shown in Figure~\ref{fig:LminR}, where the approximation values are depicted by dashed lines.
In Figure~\ref{fig:LminR}, the bounds on false positive rate and false negative rate are considered equal. 
Similar to Section~\ref{subsec:christWORandomChunk}, as the average conditional entropy per token for the generated text increases, lower number of tokens are required to achieve the same false negative and positive rates.
Comparing the results in Figures~\ref{fig:Lmin} and~\ref{fig:LminR} shows, having initial random chunk $R$ and the embedding message set $\mathcal{M}$ in DISC, increases the required number of watermarked tokens by a factor about $1.6$. 
\begin{figure}[t]
\centering
\includegraphics[scale = 0.45]{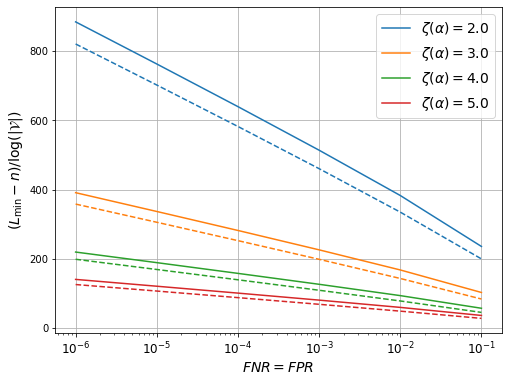}
\caption{Exact and approximate (dashed lines) $L_{\min}$ for $|\mathcal{V}|= 50272$, $n = 3h\lceil\log|\mathcal{V}|\rceil$, $|\mathcal{M}|=2^{10}$.}
\label{fig:LminDSIC}
\end{figure}

\end{document}